\documentclass[sigconf]{acmart}
\AtBeginDocument{%
  }

\copyrightyear{2025}
\acmYear{2025}
\setcopyright{cc}
\setcctype{by}
\acmConference[CIKM '25]{Proceedings of the 34th ACM International Conference on Information and Knowledge Management}{November 10--14, 2025}{Seoul, Republic of Korea}
\acmBooktitle{Proceedings of the 34th ACM International Conference on Information and Knowledge Management (CIKM '25), November 10--14, 2025, Seoul, Republic of Korea}\acmDOI{10.1145/3746252.3761064}
\acmISBN{979-8-4007-2040-6/2025/11}

\usepackage{cleveref}
\usepackage{multirow}
\usepackage{subcaption}
\usepackage{algorithm}
\usepackage{algpseudocode}

\begin{document}

\title{SupLID: Geometrical Guidance for Out-of-Distribution Detection in Semantic Segmentation}
\author{Nimeshika Udayangani}
\affiliation{%
 \institution{The University of Melbourne}
 \city{Melbourne}
 \country{Australia}}
\email{nhewadehigah@student.unimelb.edu.au}

\author{Sarah Erfani}
\affiliation{%
 \institution{The University of Melbourne}
 \city{Melbourne}
 \country{Australia}}
\email{sarah.erfani@unimelb.edu.au}

\author{Christopher Leckie}
\affiliation{%
 \institution{The University of Melbourne}
 \city{Melbourne}
 \country{Australia}}
\email{caleckie@unimelb.edu.au}


\begin{abstract}

Out-of-Distribution (OOD) detection in semantic segmentation aims to localize anomalous regions at the pixel level, advancing beyond traditional image-level OOD techniques to better suit real-world applications such as autonomous driving.
Recent literature has successfully explored the adaptation of commonly used image-level OOD methods—primarily based on classifier-derived confidence scores (e.g., energy or entropy)—for this pixel-precise task.
However, these methods inherit a set of limitations, including vulnerability to overconfidence.
In this work, we introduce SupLID, a novel framework that effectively guides classifier-derived OOD scores by exploiting the geometrical structure of the underlying semantic space, particularly using Linear Intrinsic Dimensionality (LID).
While LID effectively characterizes the local structure of high-dimensional data by analyzing distance distributions, its direct application at the pixel level remains challenging.
To overcome this, SupLID constructs a geometrical coreset that captures the intrinsic structure of the in-distribution (ID) subspace.
It then computes OOD scores at the superpixel level, enabling both efficient real-time inference and improved spatial smoothness.
We demonstrate that geometrical cues derived from SupLID serve as a complementary signal to traditional classifier confidence, enhancing the model's ability to detect diverse OOD scenarios.
Designed as a post-hoc scoring method, SupLID can be seamlessly integrated with any semantic segmentation classifier at deployment time.
Our results demonstrate that SupLID significantly enhances existing classifier-based OOD scores, achieving state-of-the-art performance across key evaluation metrics, including AUR, FPR, and AUP.
Code is available at \url{https://github.com/hdnugit/SupLID}.
\end{abstract}

\ccsdesc[500]{Computing methodologies~Scene anomaly detection}
\ccsdesc[500]{Computing methodologies~Image segmentation}

\keywords{Out-of-Distribution (OOD) Detection, Semantic Segmentation, Local Intrinsic Dimensionality (LID)}



\maketitle

\section{Introduction}
\label{sec:intro}

\begin{figure*}
    \centering
    \setlength{\tabcolsep}{0pt}
    \renewcommand{\arraystretch}{0} 
    \begin{tabular}{ccccc}
        \includegraphics[width=0.20\textwidth]{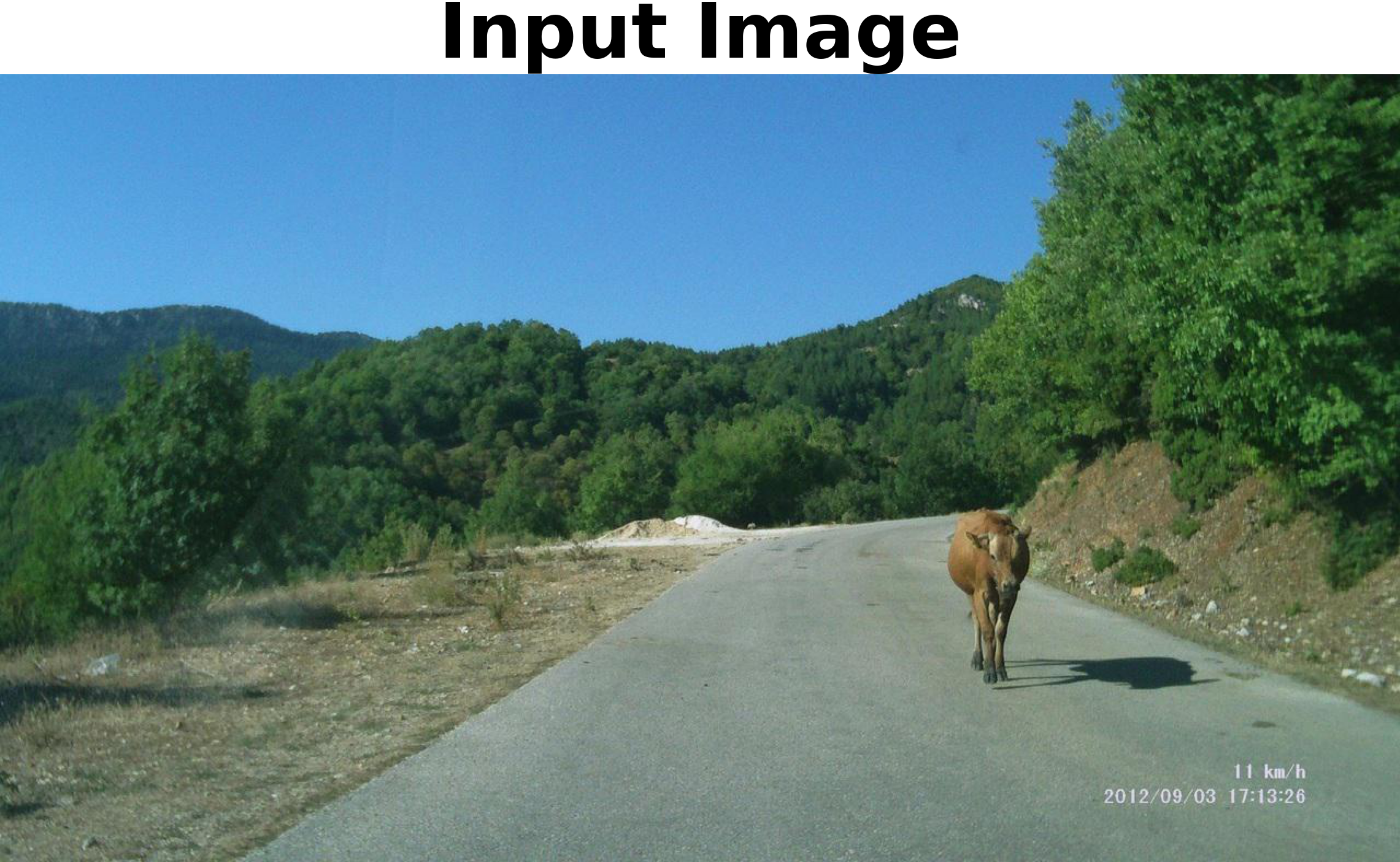} &
        \includegraphics[width=0.20\textwidth]{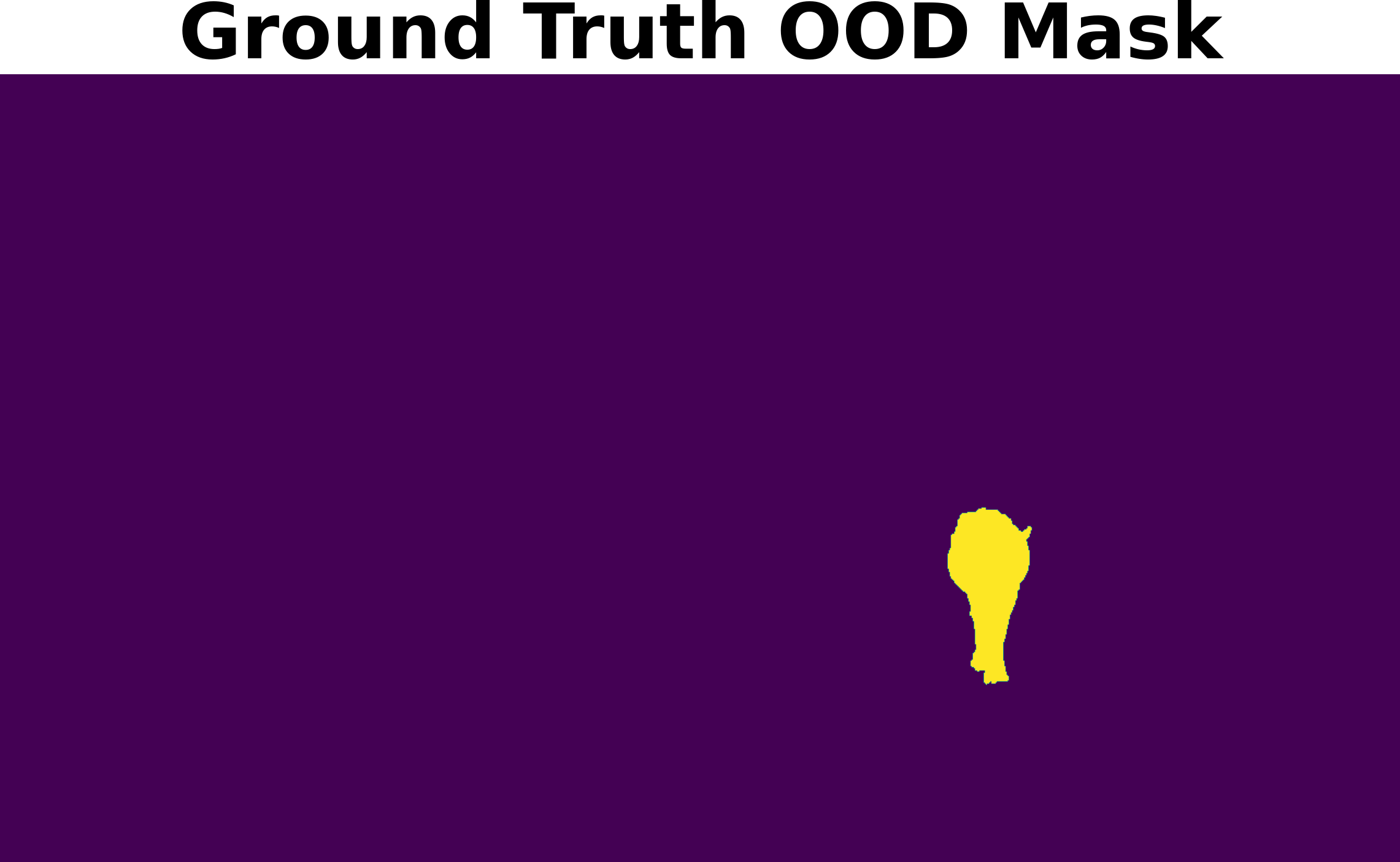} &
        \includegraphics[width=0.20\textwidth]{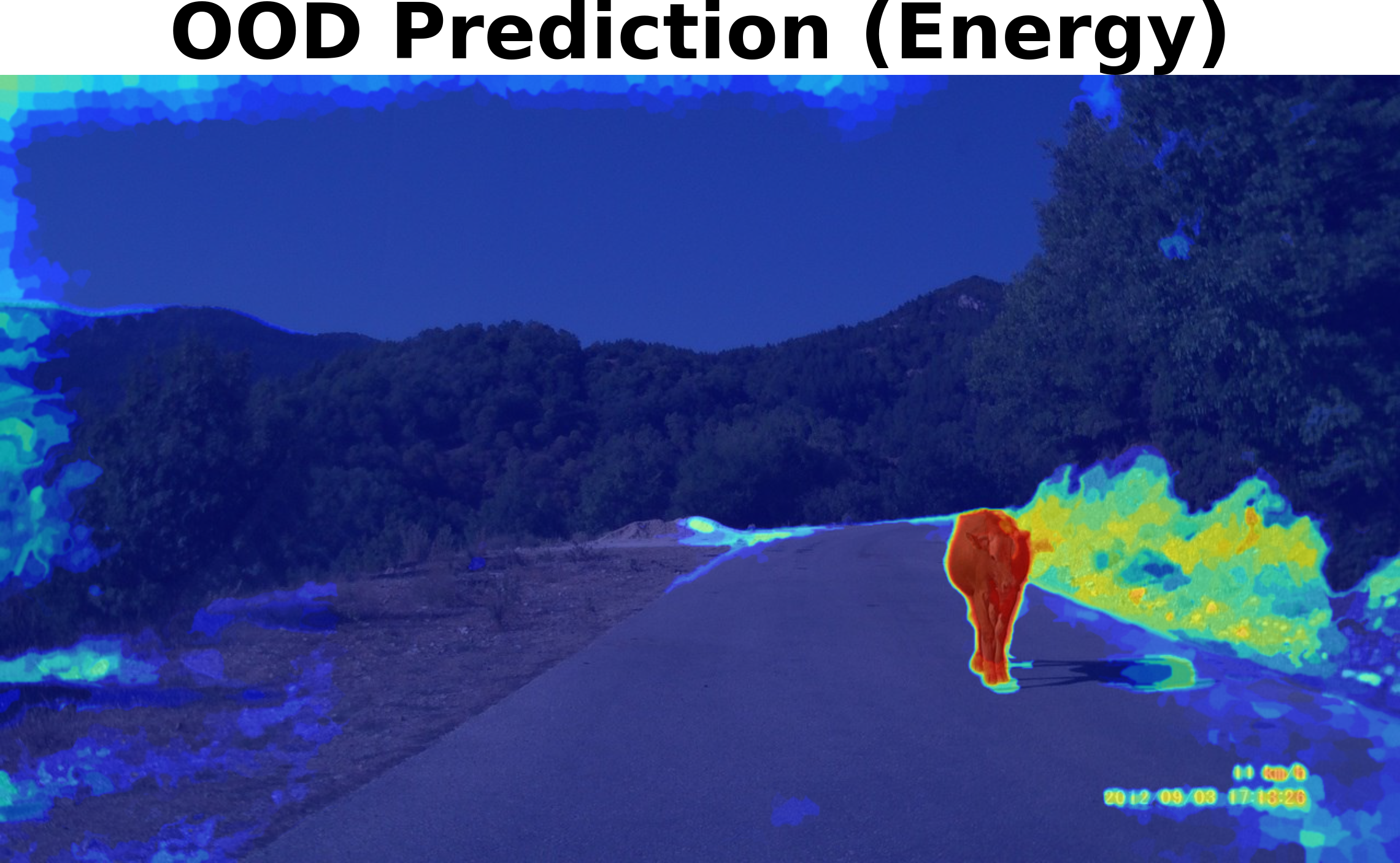} &
        \includegraphics[width=0.20\textwidth]{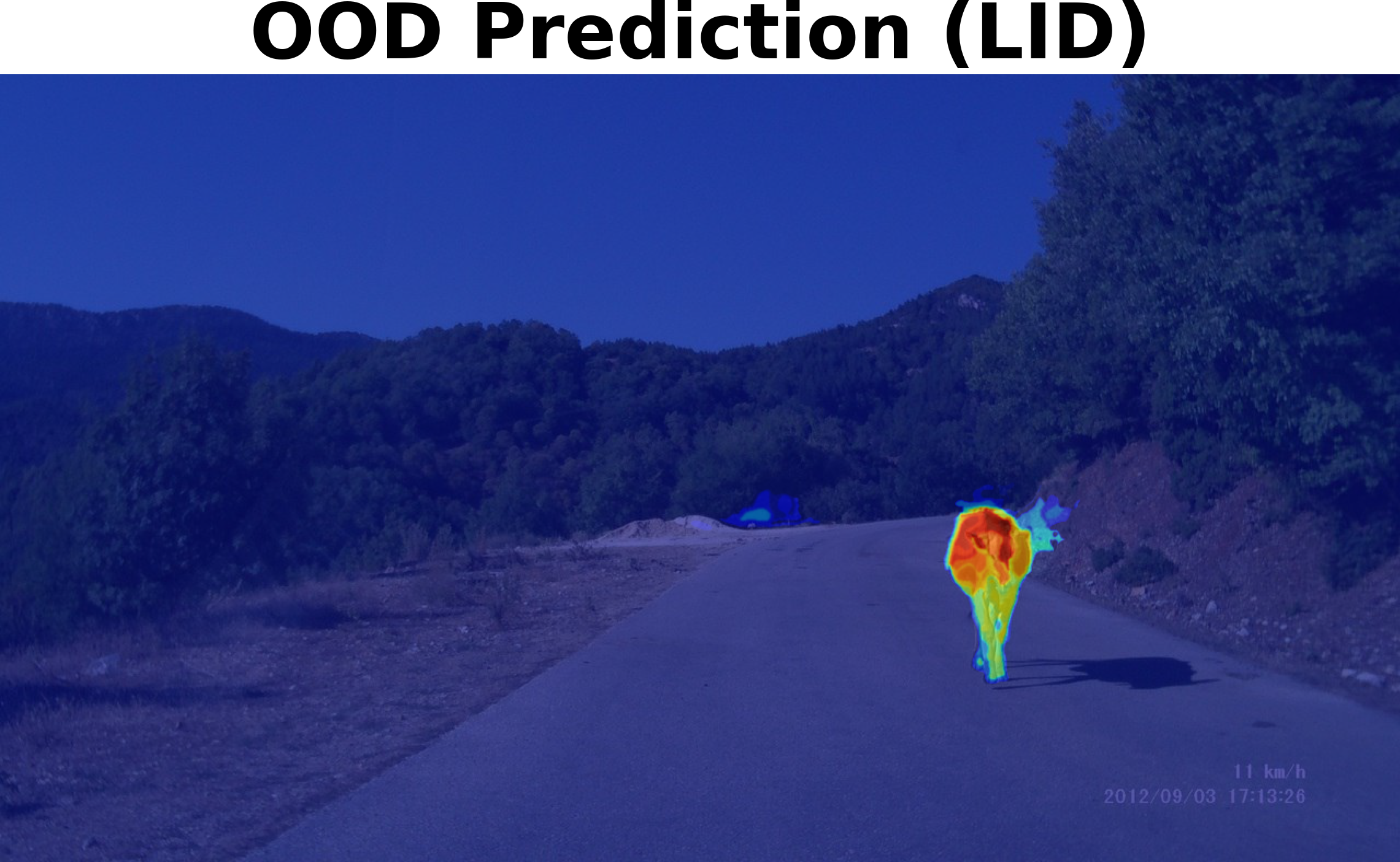} &
        \includegraphics[width=0.20\textwidth]{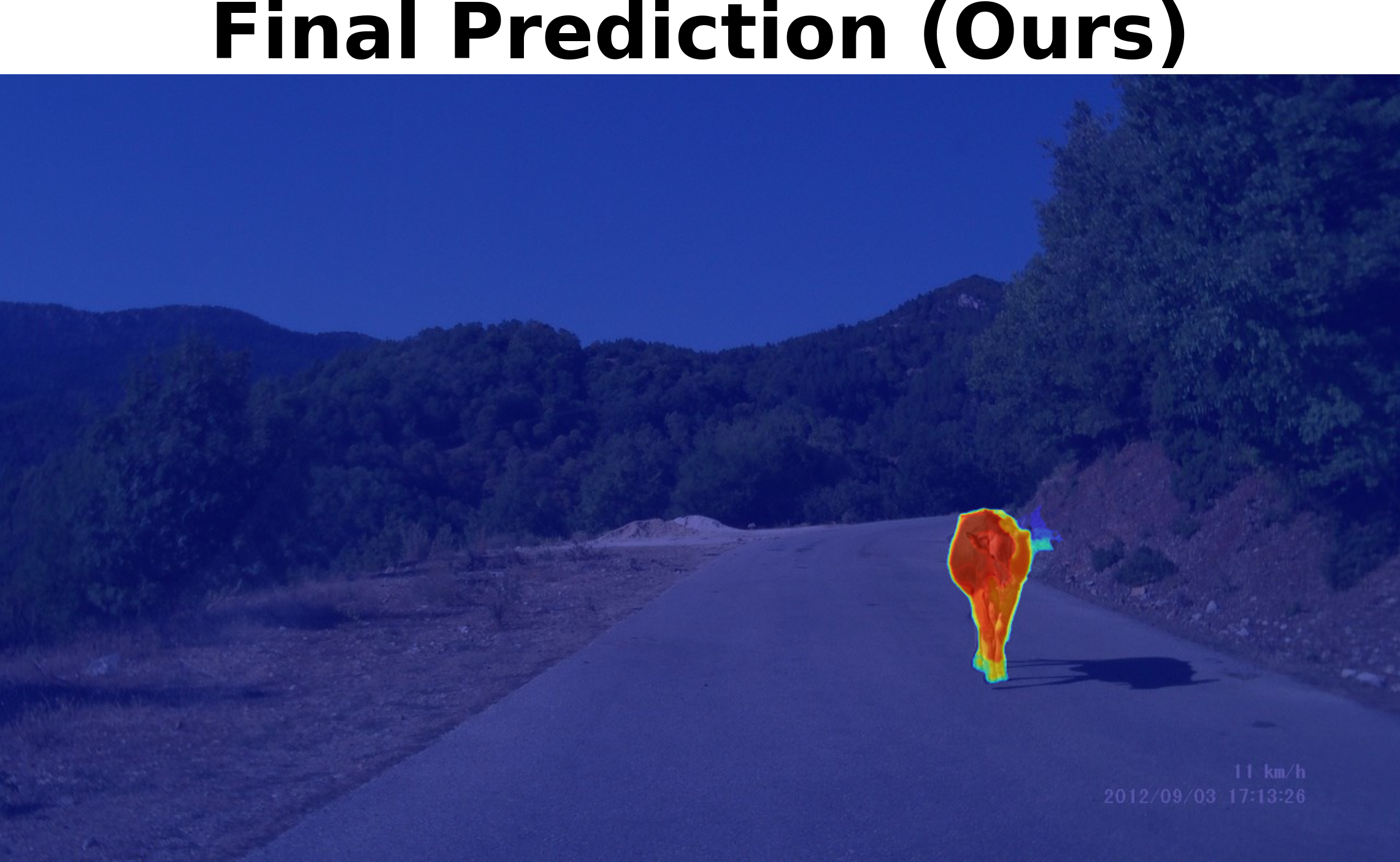} \\     
        \includegraphics[width=0.20\textwidth]{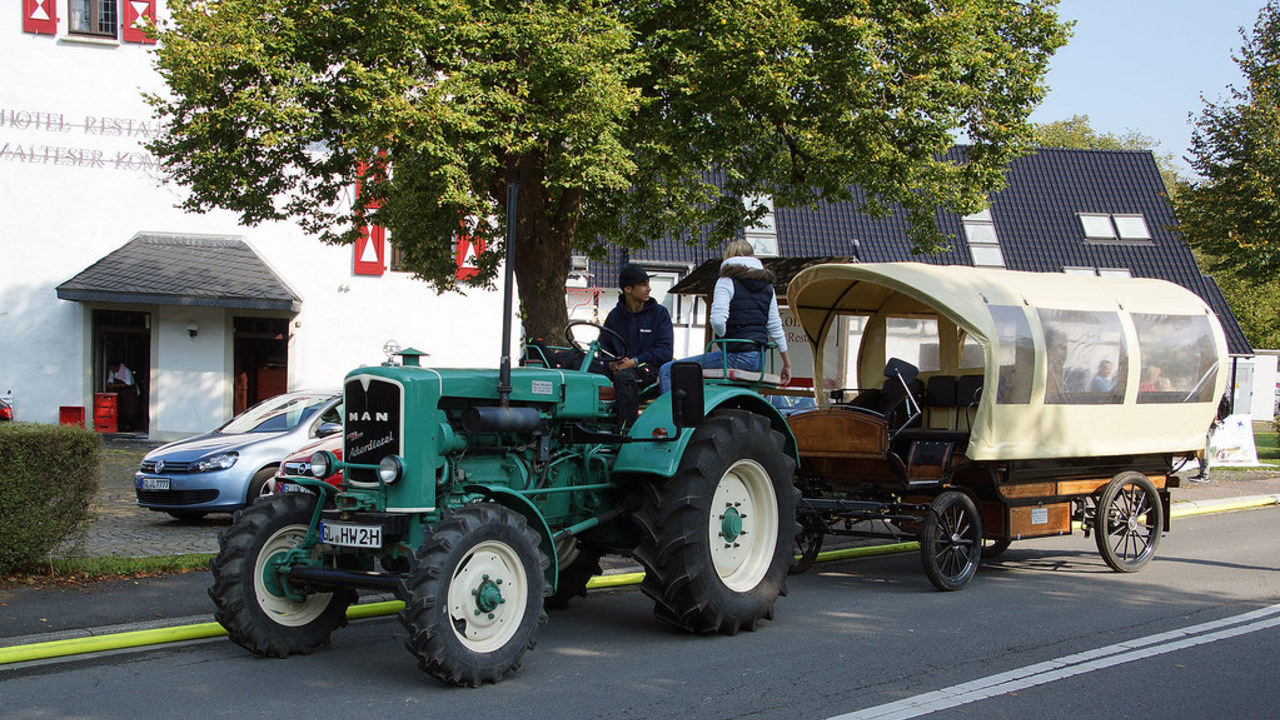} &
        \includegraphics[width=0.20\textwidth]{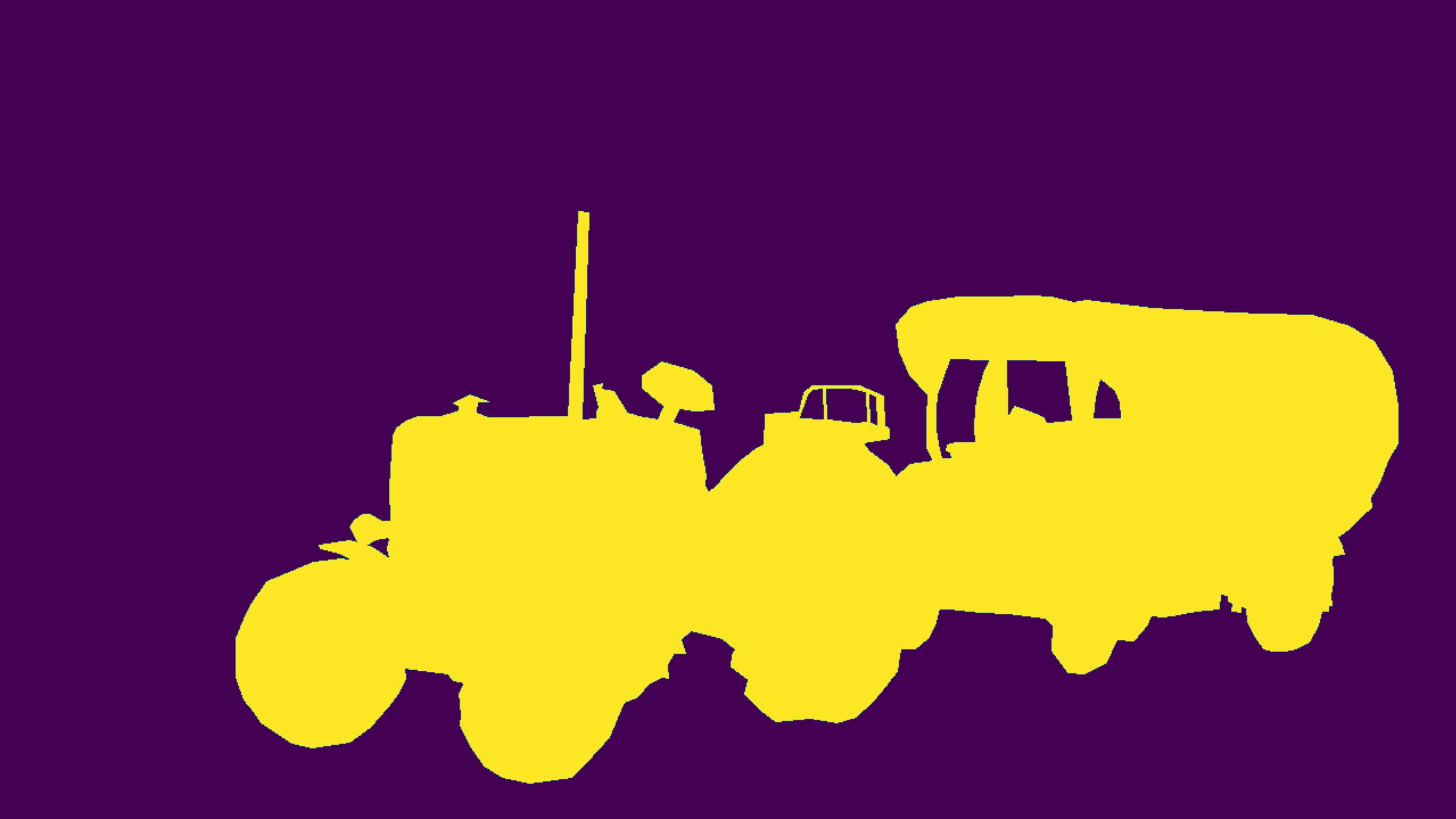} &
        \includegraphics[width=0.20\textwidth]{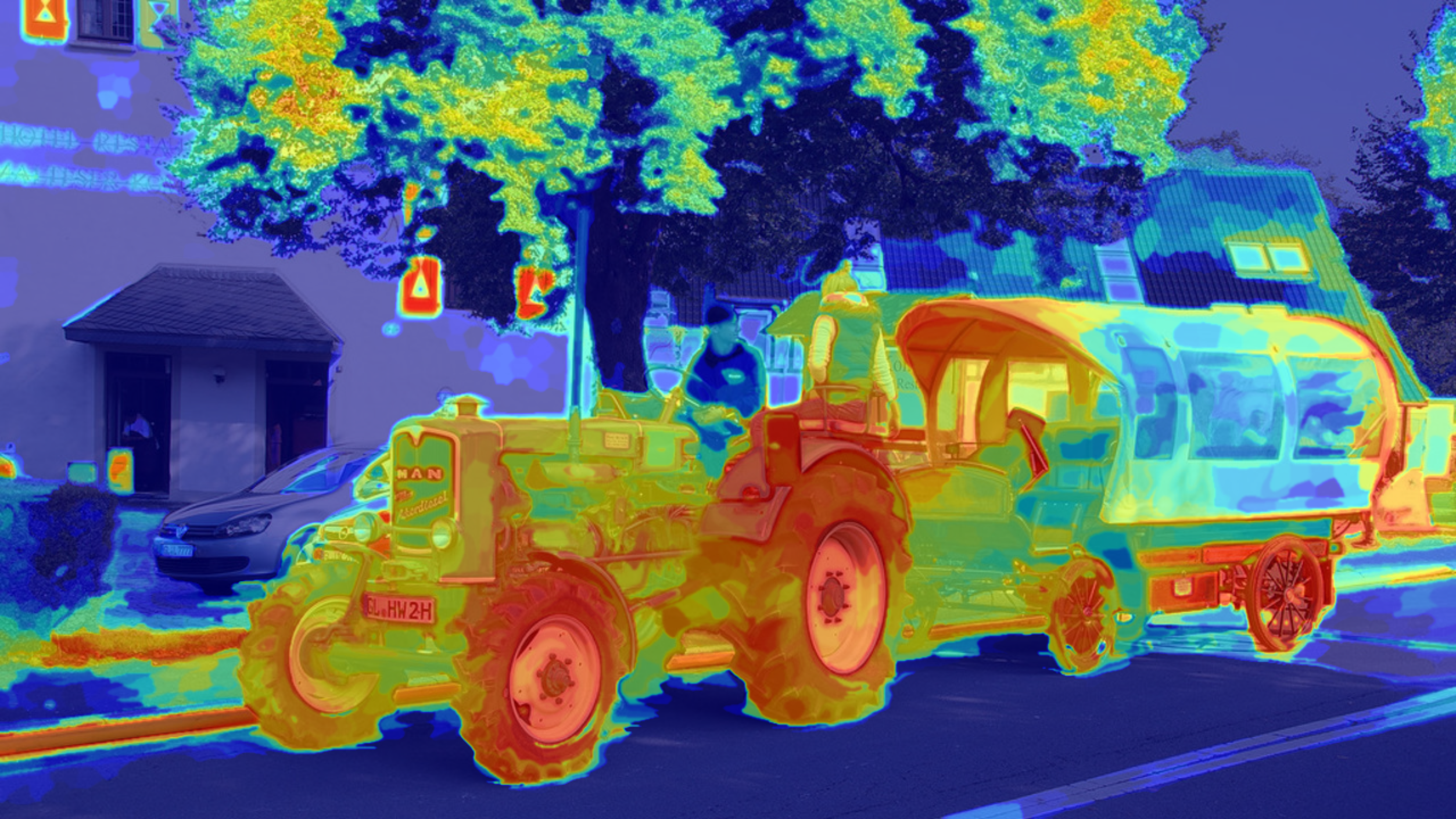} &
        \includegraphics[width=0.20\textwidth]{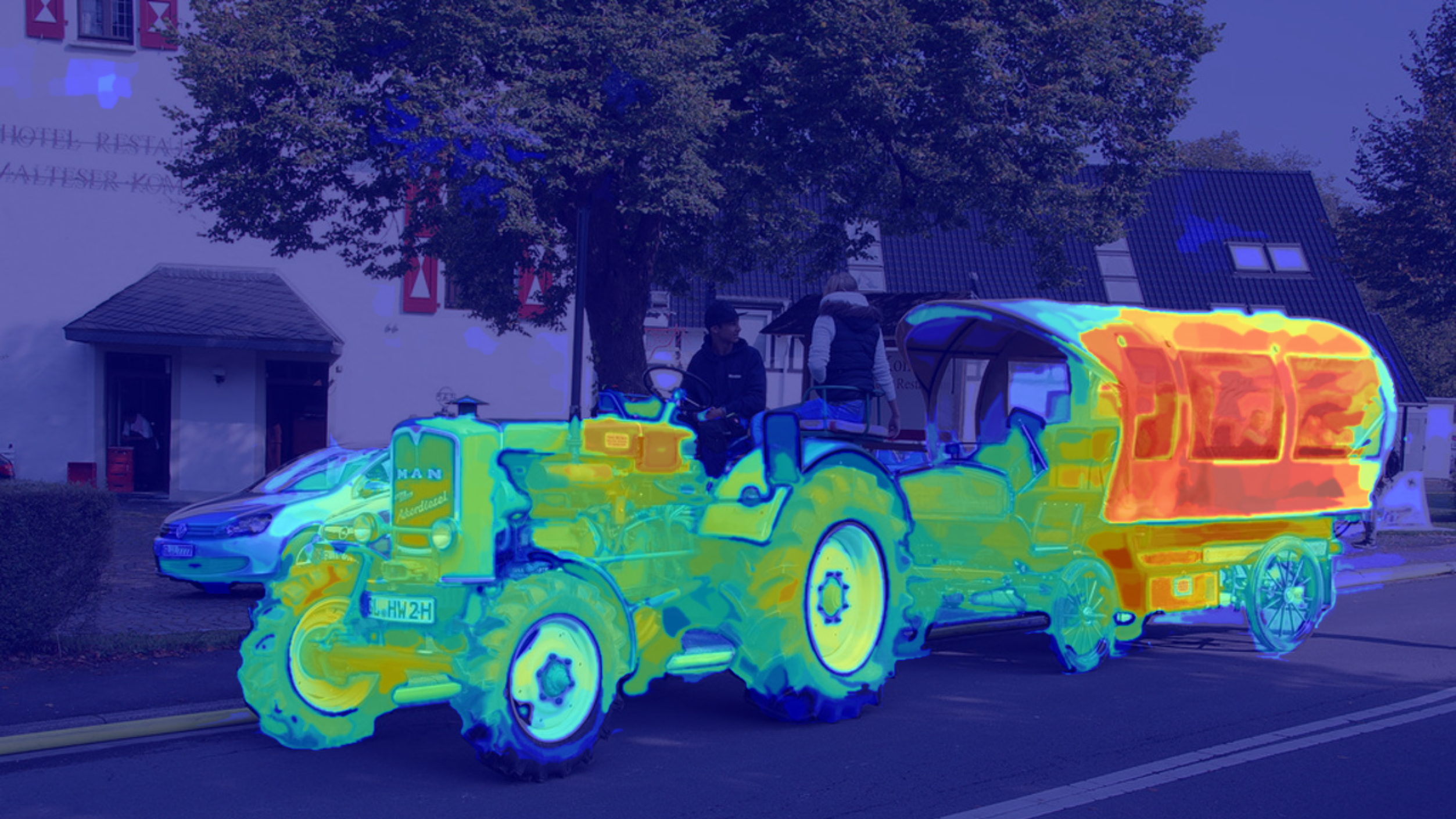} &
        \includegraphics[width=0.20\textwidth]{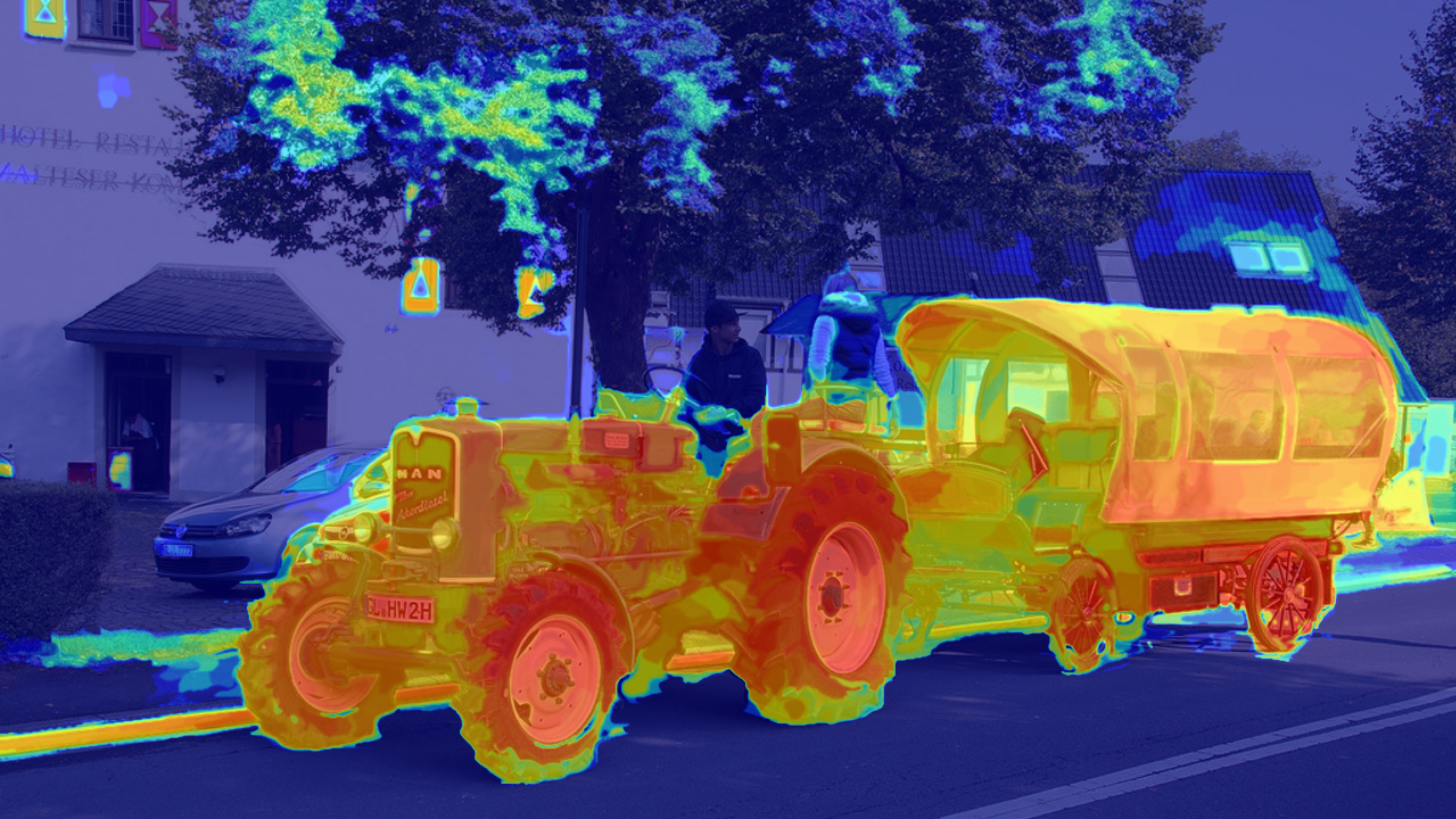}
    \end{tabular}
    \caption{Pixel-wise OOD scores overview. Example OOD score maps are shown for a SOTA segmentation model (DeepLabv3+ by NVIDIA~\cite{DeepLab_NVIDIA}) and a re-trained model for OOD detection (PEBAL~\cite{PEBAL}) using a classifier-based score (energy), followed by OOD scores generated using LID based on the proposed coreset. The two methods exhibit complementary properties: LID assigns lower scores to false positive regions and signals missed anomaly regions, but may fail to capture full anomalous objects due to its local nature. Our method (last column) produces more robust and complete OOD maps compared to both approaches. }
    \label{fig:teaser}
\end{figure*}

Out-of-distribution (OOD) detection aims to reliably identify samples that are different from the training distribution, a crucial requirement for deploying deep neural networks in open-world.
While many studies focus on detecting entire images as OOD in the context of image classification, this coarse-grained approach is insufficient for safety-critical applications such as autonomous driving. 
For example, road scenes predominantly contain in-distribution (ID) pixels, while unusual objects that are not part of the training data (OOD) may occupy only a small fraction of the image.
To address this limitation, a recent line of research~\cite{PEBAL, MetaOOD, choiBalancedEnOE, RPL, synboost} has emerged to assign per-pixel OOD scores, which can be grouped under OOD detection in the semantic segmentation task.

Many such methods extend classical classifier-based confidence scores originally proposed for image-wise OOD detection to semantic segmentation networks.
For example, MetaOOD~\cite{MetaOOD} extends the entropy score~\cite{shannonEntropy}, while MaxLogits~\cite{MaxLogit} and standardized MaxLogits~\cite{standardMaxLogit} utilize logits for anomaly pixel localization. Similarly, RPL~\cite{RPL} and PEBAL~\cite{PEBAL} employ energy score~\cite{EnergyOE}, which also leverage pre-softmax activations, as pixel-wise OOD scores. However, these methods have certain limitations.
While they are effective at identifying OOD samples that are closer to classification decision boundaries, corresponding to low-confidence regions~(near-OOD), they can be overconfident on far-OOD samples, often misclassifying them as ID when they are distant from the ID region~\cite{NNGuide, PEBAL}.
Furthermore, there is no guarantee that all samples near the classification boundaries are OOD, as some challenging ID samples may also reside in these regions.
Hence, these methods can result in a large number of false positive and false negative OOD detections~\cite{PEBAL}.

To mitigate these limitations, we propose exploiting geometrical properties of the data's semantic space—particularly Linear Intrinsic Dimensionality (LID)—as a complementary signal for characterizing outlierness.
LID captures the local structure of high-dimensional data based on the distance distribution of a reference data sample to its neighbors~\cite{LID_adversarial}, representing the dimensionality of the data subspace local to that point.
Empirically, high LID values are associated with samples that are likely to lie outside a given data subspace~\cite{LID_adversarial, LID_OOD_correlation}.
Hence, we argue that this dimensional insight can be effectively leveraged to guide decisions on challenging ID and OOD samples in the semantic space.
Also being a non-parametric estimate, LID does not rely on any classifier-specific parameters, and thus does not inherit the limitations associated being solely relying on classifier-based scores.
Our experiments show that LID can effectively compensate for missing information in classifier-based scores.
For example, in~\Cref{fig:teaser}, the OOD score derived from LID (column 4) assigns lower scores to false positive regions and highlights anomaly regions that are missed by the energy-based OOD score alone (column 3).



The empirical relationship between LID and outlierness has already been established~\cite{LID_OOD_correlation,LID_adversarial}. It has also been successfully applied to characterize adversarial subspaces~\cite{LID_adversarial}, mitigate the effect of noisy labels~\cite{LID_noisy_label}, and backdoor attacks~\cite{COLLIDER_LID_coreset}.
Although LID has been used for such instance-level tasks~\cite{LID_reconstruction}, it is a significant challenge to leverage LID in the pixel-wise anomaly segmentation tasks particularly due to the computational complexity associated with LID calculation, as it involves estimating neighbor distance distributions.
To address this, we propose a novel framework, \textit{SupLID}, which develops two key techniques: \textit{superpixel-based aggregation} and \textit{geometrical coreset selection}.
The first technique involves calculating OOD scores at a lower spatial resolution by grouping pixels based on local and intensity information in a class-agnostic manner.
This not only reduces the computational complexity required for real-time anomaly localization, but also improves spatial smoothness (improving the ability to classify neighboring object pixels~\cite{PEBAL}), and reduces fragmentation along object boundaries.
Secondly, we propose a novel geometrical coreset to capture the intrinsic structure of the ID subspace.
Coreset selection~\cite{coreset_survey}, in applications with massive data, attempts to find a weighted subset that serves as a compact representation of the whole data.
We utilize LID to form a geometrical coreset that serves as a reference neighborhood subspace for better distinguishing OOD samples.
In particular, this reduces the number of neighborhood queries required for LID estimation to the order of $10^2$, enabling scalable pixel-wise OOD detection. \looseness=-1


While LID provides valuable insights into the underlying geometrical structure, it is a localized measure~\cite{LID_reconstruction} (see~\Cref{fig:teaser}, column 4).
Also, using LID as the sole criterion for OOD detection disregards the information encoded in classification weights with class-dependent signals.
Therefore, we propose utilizing LID to guide the classifier-based score, leveraging the complementary strengths of both classifier-based knowledge and the geometrical properties of the semantic space, allowing more diverse and robust OOD detection in semantic segmentation—explored for the first time in this context.
Unlike recent SOTA methods that require complex re-training~\cite{PEBAL, RPL, MetaOOD} or hybrid approaches~\cite{synboost, densehybrid} that also require an extra reconstruction network that is hard to train and inefficient to run in real-time, our method requires no retraining.
It can be effortlessly applied to any semantic segmentation network at deployment, serving as a post-hoc technique to enhance classifier-derived detection performance.
In summary, our contributions are as follows:
\begin{itemize}
\item We propose a novel method for OOD detection in semantic segmentation, SupLID, which accounts for local variations in intrinsic dimensionality within the data to enhance classifier-based OOD scores.
SupLID mutually reinforces classifier-based knowledge and geometrical structure of the semantic space, without requiring network retraining, and thus does not compromise the segmentation of ID pixels.
\item We introduce novel geometrical coreset selection and super\-pixel-based aggregation to enable scalable pixel-wise inference and enhance smoothness properties.
\item We demonstrate the effectiveness of our method on diverse OOD benchmarks— RoadAnomaly~\cite{RoadAnomaly_dataset}, SMIYC~\cite{SMIYC_dataset}, and Fishyscapes~\cite{fishyscapes}— and the generality of SupLID across a broad range of classifier-based scores.
\end{itemize}

\section{Related Work}\label{sec:Related_work}

The most common and primary approach to OOD detection is to leverage a confidence score derived from deep neural classifiers to identify OOD samples, such as the maximum of softmax probability (MSP)~\cite{hendrycks_2017MSP}, entropy~\cite{shannonEntropy}, energy~\cite{EnergyOE}—which also mirrors the class conditional probability—Kullback–Leibler (KL)~\cite{MaxLogit} divergence, and the maximum of logit (MaxLogit)~\cite{MaxLogit, Maxlogit_openset}. 
Initially, these metrics were evaluated for image-level OOD detection but have since been extended to pixel-wise OOD detection in semantic segmentation.
For example, \cite{MaxLogit, standardMaxLogit} explore MSP and MaxLogit to evaluate OOD detection in semantic segmentation. PEBAL~\cite{PEBAL} utilizes energy as the pixel-wise OOD score, combined with pixel-level abstention class learning. 
MetaOOD~\cite{MetaOOD} investigates the entropy score and applies the traditional Outlier Exposure~\cite{hendrycks_2018OE} method to semantic segmentation networks, while BE-OE~\cite{choiBalancedEnOE} addresses the class imbalance issue associated with Outlier Exposure.
RPL~\cite{RPL} also incorporates the energy score and introduces context-robust contrastive learning losses to improve the generalization of OOD pixel detection to new contexts.

In contrast to these classifier-based scores, non-parametric OOD detection methods such as kNN~\cite{KNN_OOD}, Mahalanobis~\cite{lee_2018_mahalanobis} and density estimation~\cite{embeddingDensity} provide greater flexibility, as they do not rely on network parameters~\cite{DAO_LID_outlier}.
Since these approaches are based on the distance to ID data or local density in feature space, they can reliably assign low scores to far-OOD regions~\cite{NNGuide}. However, they have been rarely explored for pixel-wise OOD detection.

Whether classifier-based or non-parametric, their performance is limited by relying on a single source of information—either disregarding the knowledge embedded in classification weights or ignoring distance-based information in the feature space~\cite{NNGuide}.
NNGuide~\cite{NNGuide} is the only method that attempts to combine the complementary strengths of these two approaches, but it is designed for image-level OOD detection rather than pixel-wise detection.
They propose to use kNN distance to guide the OOD score derived from the classifier-based confidence score, aiming to reduce the overconfidence of OOD samples by ensuring that the classifier-based score respects the boundary geometry of the data manifold.
However, like kNN, most non-parametric methods rely on the distances between a test point and its nearest neighbors. As a result, these methods are affected by the well-known "curse of dimensionality" where the quality of distance-based information diminishes as the dimensionality of the data increases~\cite{DAO_LID_outlier}.
Moreover, they do not explicitly utilize the underlying geometrical structure of the data.
In contrast, such high-dimensional data can be more effectively characterized using Local Intrinsic Dimensionality~(LID).
Also, unlike Mahalanobis distance, LID makes no explicit assumptions about the underlying data distribution.

LID has been successfully applied to identify adversarial subspaces~\cite{LID_adversarial} and backdoor-poisoned data~\cite{COLLIDER_LID_coreset}.
In addition, \citet{LID_reconstruction} use LID to improve OOD detection in reconstruction networks.
However, its application in pixel-wise tasks has not been explored primarily due to the computational cost.

It is also worth noting that a separate line of recent work (e.g., SynBoost~\cite{synboost}, DenseHybrid~\cite{densehybrid}) on OOD detection in semantic segmentation has proposed combining information from multiple sources, similar in spirit to our approach.
These methods suggest leveraging both classifier-based and generative-based knowledge.
However, as hybrid approaches, they typically require an additional network that is hard to train and inefficient for real-time inference.
Without retraining a separate network or adding a new set of layers, these methods cannot function as post-hoc solutions. 
 Also, none of these approaches utilize geometrical structure underlying semantic space.
This is the first work to explore the untapped potential of geometrical properties to enhance classifier-based OOD detection in semantic segmentation, operating in a fully post-hoc manner.

\section{\hspace{-0.5em}Background on Local Intrinsic Dimensionality\hspace{-0.5em}} \label{sec:background}

Local Intrinsic Dimensionality (LID) can be intuitively understood as the dimensionality of the data subspace formed by local samples around a given data point~\cite{DAO_LID_outlier}.
LID, similar to classical expansion models~\cite{classicEx,GED}, is based on the observation that the volume $V$ of a $d$-dimensional ball grows proportionally to $r^d$ as its radius is scaled by a factor of $r$, which yields the dimension $d$ as:
\begin{equation*}
\frac{V_2}{V_1} = \left( \frac{r_2}{r_1} \right)^d \quad \Rightarrow \quad d = \frac{\ln(V_2/V_1)}{\ln(r_2/r_1)},
\end{equation*}
where $V_1$ and $V_2$ represent the volumes of two concentric balls with radii $r_1$ and $r_2$.
To estimate $d$, these methods approximate the volume by the number of data points contained within the ball.
In contrast, LID~\cite{LID,DAO_LID_outlier} extends this discrete notion into a continuous approximation by representing the volume of a ball with the probability measure associated with its interior.

\begin{definition}[Local Intrinsic Dimensionality~\cite{LID}]
Given a data sample $\boldsymbol{x} \in \mathcal{X}$, let $r>0$ denote a non-negative random variable that measures the distance of $\boldsymbol{x}$ to other data samples, and assume its cumulative distribution function is denoted by $F(r)$. If $F(r)$ is positive and continuously differentiable for every $r>0$, then the intrinsic dimensionality of $\boldsymbol{x}$ at distance $r$ is given by
\begin{equation*}
\mathrm{ID}_F(r) \triangleq \lim _{\epsilon \rightarrow 0^{+}} \frac{\ln (F((1+\epsilon) r) / F(r))}{\ln (1+\epsilon)}=\frac{r F'(r)}{F(r)},
\end{equation*}
whenever the limit exists. The LID at $\boldsymbol{x}$ can then be defined as the limit of $\mathrm{ID}_F(r)$ as $r \rightarrow 0^{+}$ :
\begin{equation}\label{eq:LID}
\mathrm{LID}_F=\lim _{r \rightarrow 0^{+}} \mathrm{ID}_F(r).
\end{equation}
\end{definition}
Solving \Cref{eq:LID} is not straightforward, as it requires knowledge of the exact distance distribution $F(r)$. In~\cite{estimatingLID}, a framework based on extreme value theory was used to derive several estimators of $\mathrm{LID}_F$, among which we chose the maximum likelihood estimator (MLE) due to its efficacy~\cite{LID_OOD_correlation, estimatingLID}.
\begin{equation}\label{eq:LIDMLE}
\widehat{\mathrm{LID}}(\boldsymbol{x})=-\left(\frac{1}{k} \sum_{i=1}^k \log \frac{r_i(\boldsymbol{x})}{r_k(\boldsymbol{x})}\right)^{-1},
\end{equation}
where $r_i(\boldsymbol{x})$ denotes the distance from $\boldsymbol{x}$ to its $i$-th nearest neighbor, and thus $r_k(\boldsymbol{x})$ is the maximum of the neighbor distances.
As seen, this model requires no explicit knowledge of the underlying data distribution and ultimately reveals relationships among the expected sizes and radii of neighborhoods around points of interest—the same information used in many state-of-the-art distance-based OOD detection methods~\cite{LID_OOD_correlation}.
However, unlike these methods, it is naturally adaptive to local variations in intrinsic dimensionality, allowing it to work well even when the dimensional characteristics are highly non-uniform~\cite{LID_OOD_correlation}.

\section{Methodology} \label{sec:method}
In this section, we present our SupLID, beginning with a formal introduction to the OOD detection problem in semantic segmentation.
Then, we describe the SupLID framework, which incorporates superpixel-based aggregation and geometrical coreset selection to leverage the geometrical properties of the semantic space, followed by the definition of the novel post-hoc SupLID OOD score.

\subsection{Problem Definition}
Let $\mathcal{X} \subset \mathbb{R}^{H \times W \times C}$ be the space of input images of height $H$, width $W$, and $C$ channels, and let $\mathcal{Y}=\{1,2, \ldots, K\}$ be the set of in-distribution $\left(\mathcal{D}^{\text {in }}\right)$ semantic classes. A segmentation model $f:$ $\mathcal{X} \rightarrow \mathbb{R}^{H \times W \times K}$, typically trained on samples from $\mathcal{D}_{\text {in }}$, maps an input image $\boldsymbol{x} \in \mathcal{X}$ to a per-pixel distribution over $K$ classes.

The pixel-wise OOD detection problem in semantic segmentation aims to define a scoring function $S: \mathcal{X} \rightarrow \mathbb{R}^{H \times W}$ that assigns to each pixel at location $(i, j)$ of $\boldsymbol{x}$ ($x_{i, j}$),  a score denoted as $S_{i, j}(\boldsymbol{x})$. The goal is for the function $S$ to assign a high score $S_{i, j}(\boldsymbol{x})$ for pixels $x_{i, j} \sim \mathcal{D}_{\text {in }}$ and a low score for out-of-distribution pixels, i.e., pixels that depict content not seen during training or do not belong to any class in $\mathcal{Y}$.

A threshold $\tau$ is then used to classify each pixel:
$$
\operatorname{Decision}\left(x_{i, j}\right)= \begin{cases}\text { ID } & \text { if } S_{i, j}(\boldsymbol{x}) \geq \tau \\ \mathrm{OOD} & \text { if } S_{i, j}(\boldsymbol{x})<\tau\end{cases}
$$

\captionsetup[algorithm]{font=footnotesize}  
{\footnotesize
\begin{algorithm}[t]
\caption{LID-Based Geometrical Coreset Selection}
\begin{algorithmic}[1]
\Require Superpixel embeddings $\{\mathcal{Z}_y\}_{y \in \mathcal{Y}}$ per class, coreset size $m$
\Ensure Coreset $\mathcal{Z}'$ and associated weights $\mathcal{W}'$

\State Initialize $\mathcal{Z}' \gets \emptyset$, $\mathcal{W}' \gets \emptyset$
\For{each class label $y \in \mathcal{Y}$}
    \State Let $\mathcal{Z}_y = \{\boldsymbol{z}_1, \boldsymbol{z}_2, \dots, \boldsymbol{z}_n\}$ be set of superpixel embeddings for class $y$
    \State Compute $\widehat{\mathrm{LID}}(\boldsymbol{z}_i)$ for all $\boldsymbol{z}_i \in \mathcal{Z}_y$
    \State Select subset $\mathcal{Z}_y' \subset \mathcal{Z}_y$ with $|\mathcal{Z}_y'| = m$ having lowest total LID:
    \[
        \mathcal{Z}_y' = \underset{T \subseteq\mathcal{Z}_y, |T| = m}{\arg\min} \sum_{\boldsymbol{z}_t \in T} \widehat{\mathrm{LID}}(\boldsymbol{z}_t)
    \]
    \State Let $\mathcal{W}_y' = \{\widehat{\mathrm{LID}}(\boldsymbol{z}_t) \mid \boldsymbol{z}_t \in \mathcal{Z}_y'\}$
    \State Update $\mathcal{Z}' \gets \mathcal{Z}' \cup \mathcal{Z}_y'$, $\mathcal{W}' \gets \mathcal{W}' \cup \mathcal{W}_y'$
\EndFor
\State \Return $\mathcal{Z}'$, $\mathcal{W}'$
\end{algorithmic}
  \label{algo:coreset}
\end{algorithm}
}

\subsection{Proposed SupLID}
Let \( S: \mathcal{X} \rightarrow [0, \infty)^{H \times W} \) be a classifier-based confidence score function for OOD detection such as energy~\cite{EnergyOE}.
As discussed in~\Cref{sec:intro}, such a score may not be sufficient for reliable OOD detection, particularly due to the overconfidence issue.
Therefore, we suggest to utilize the geometrical structure of the underlying data to further improve this score.

Following~\cite{NNGuide}, a distance-based guidance can be introduced to \( S \) as,
$$
S^{Guided}(\boldsymbol{x})=S(\boldsymbol{x}) \cdot D(\boldsymbol{x}),
$$
for $\boldsymbol{x} \in \mathcal{X}$, where $D(\boldsymbol{x})$ represents the distance-based guidance score.

However, traditional distance-based methods like kNN are not sufficient, as they ignore dimensionality and geometrical structure of the underlying data. It has been shown that such methods are susceptible to variations in intrinsic dimensionality~\cite{DAO_LID_outlier}. In contrast, LID offers valuable insight into outlierness, especially for datasets where LID values vary significantly, such as in cases with substantial domain or context differences between training and testing semantic spaces.
Therefore, $D(\boldsymbol{x})$ is chosen to be derived from LID.

However, computing the LID estimate in~\Cref{eq:LIDMLE} requires kNN queries, which are computationally prohibitive in this pixel-wise setting—unlike in instance-wise scenarios—particularly when using a large ID training set as the reference and dealing with high-resolution images, which can result in pixel counts on the order of millions.
To address this, we introduce two key steps in our method: superpixel-based aggregation and geometrical coreset selection.

\subsubsection{Superpixel-based Aggregation}
We propose calculating the $S^{Guided}$ score at a lower resolution by grouping image pixels into spatially coherent regions. To this end, we adopt the classical SLIC~\cite{SLIC} algorithm for superpixels, which preserves both locality and object boundaries.
This integration offers several benefits to our framework. First, it significantly reduces computational complexity by a factor of $H \times W / N$, where $N$ is the desired number of superpixels. More importantly, superpixels smooth out pixel-level noise while preserving boundary information. This helps reduce the fragmentation typically observed in current anomaly object masks, allowing hard pixels to benefit from the scores of their local neighbors.
Additionally, superpixels generate class-agnostic segments, enabling them to operate equally well on unseen anomalous pixels.

\noindent Therefore, the $S^{Guided}$ score at superpixel $\boldsymbol{x}_l $, where $l \in \{0, 1, \ldots, N\}$ of the input image $\boldsymbol{x} \in \mathcal{X}$, denoted as $S^{Guided}_l$, is obtained as: $S^{Guided}_l(\boldsymbol{x}) = S_l(\boldsymbol{x}) \cdot D_l(\boldsymbol{x})$, where,
\begin{equation} \label{eq:sl_x}
S_l(\boldsymbol{x})= \frac{1}{|\boldsymbol{x}_l|} \sum_{\forall(i, j) \text { s.t. } x_{i j,} \in \boldsymbol{x}_l} S_{i, j}(\boldsymbol{x})  
\end{equation}
is the average confidence score over all pixels belonging to the superpixel.
To proceed with the calculation of $D_l(\boldsymbol{x})$, we use the trained encoder of $f$ that maps $\boldsymbol{x}$ to an embedding $\boldsymbol{z}=\phi\left(\boldsymbol{x}\right)$, and the training set $\mathcal{X}$ to a set of neighbors $\mathcal{Z}:=\phi(\mathcal{X})$, following~\cite{LID_adversarial}. Intuitively, if $\boldsymbol{x}_l$ is $\operatorname{OOD}$, then $\boldsymbol{z}_l$ is also differently distributed; e.g. has  neighbors with different classes~\cite{fishyscapes}.

Note that even computing the estimate given in~\Cref{eq:LIDMLE} is computationally expensive, as each calculation results in $N \times |\mathcal{Z}|$ neighbor queries.

\begin{table*}[tb!]
\centering
\caption{Results comparison with post-hoc OOD scores on DeepLabv3+ WideResNet-38 semantic segmentation model. The best results are bolded. Results averaged over six random runs.}
\begin{small}
\setlength{\tabcolsep}{3.2pt}
\begin{tabular}{l*{6}{ccc}}
\toprule
\textbf{Method} & \multicolumn{3}{c}{\textbf{SMIYC-Anomaly}} & \multicolumn{3}{c}{\textbf{SMIYC-Obstacle}} & \multicolumn{3}{c}{\textbf{RoadAnomaly}} & \multicolumn{3}{c}{\textbf{FS-Static}} & \multicolumn{3}{c}{\textbf{FS-L\&F}} & \multicolumn{3}{c}{\textbf{Average}} \\
\cmidrule(lr){2-4} \cmidrule(lr){5-7} \cmidrule(lr){8-10} \cmidrule(lr){11-13} \cmidrule(lr){14-16} \cmidrule(lr){17-19}
& FPR & AUP & AUR & FPR & AUP & AUR & FPR & AUP & AUR & FPR & AUP & AUR & FPR & AUP & AUR & FPR & AUP & AUR \\
\midrule
MSP~\cite{hendrycks_2017MSP} & 60.33 & 40.81 & 78.90 & 3.86 & 41.24 & 99.10 & 71.24 & 15.79 & 67.60 & 24.33 & 22.51 & 92.16 & 40.25 & 4.96 & 89.39 & 40.00 & 25.06 & 85.43 \\
MaxLogit~\cite{MaxLogit} & 67.54 & 45.99 & 80.52 & 4.48 & 39.73 & 98.72 & 70.30 & 19.05 & 72.87 & 18.17 & 40.46 & 95.73 & 42.21 & 14.93 & 93.46 & 40.54 & 32.03 & 88.26 \\
KL~\cite{MaxLogit} & 66.77 & 45.24 & 80.68 & 4.63 & 34.51 & 98.64 & 69.65 & 19.63 & 73.51 & 17.34 & 43.20 & 96.12 & 39.78 & 18.45 & 94.08 & 39.50 & 32.21 & 88.61 \\
Entropy~\cite{MetaOOD} & 60.54 & 45.73 & 80.19 & 3.85 & 44.71 & 99.09 & 70.94 & 17.01 & 68.88 & 23.60 & 28.66 & 93.08 & 40.03 & 10.62 & 90.93 & 39.79 & 29.35 & 86.44 \\
Energy~\cite{EnergyOE} & 67.46 & 45.44 & 80.48 & 4.49 & 34.47 & 98.67 & 70.16 & 19.54 & 73.35 & 17.77 & 41.68 & 95.90 & 41.79 & 16.05 & 93.71 & 40.34 & 31.44 & 88.43 \\
Mahalanobis~\cite{lee_2018_mahalanobis} & 77.63 & 27.71 & 71.41 & 93.33 & 0.44 & 31.23 & 88.09 & 17.10 & 64.27 & 20.26 & 33.05 & 94.89 & 39.44 & 2.93 & 89.71 & 63.75 & 16.25 & 70.30 \\
kNN~\cite{KNN_OOD} & 69.41 & 39.42 & 77.15 & 63.75 & 0.95 & 65.35 & 82.24 & 16.98 & 65.31 & 35.50 & 8.28 & 87.31 & \textbf{21.60} & 2.47 & 92.65 & 54.50 & 13.62 & 77.56 \\
NNGuide~\cite{NNGuide} & 53.37 & 49.46 & 84.03 & 18.37 & 11.01 & 95.41 & 71.84 & 19.10 & 71.78 & 19.97 & 31.61 & 94.66 & 30.15 & 10.66 & \textbf{94.53} & 38.74 & 24.37 & 88.08 \\
\textbf{SupLID (Ours)} & \textbf{44.55} & \textbf{59.46} & \textbf{87.94} & \textbf{2.27} & \textbf{68.96} & \textbf{99.36} & \textbf{52.52} & \textbf{30.34} & \textbf{81.98} & \textbf{17.20} & \textbf{43.32} & \textbf{96.37} & 50.07 & \textbf{24.91} & 92.12 & \textbf{33.32} & \textbf{45.40} & \textbf{91.55} \\
\bottomrule
\end{tabular}
\end{small}
\label{tab:main_posthoc_results}
\end{table*}

\subsubsection{LID-based Coreset Selection}

To further reduce computational complexity and better capture the geometry of the in-dis\-tribution, we propose using a geometrical coreset based on Local Intrinsic Dimensionality (see \Cref{algo:coreset}).
Coreset selection involves identifying a small, weighted subset of data points that preserves the essential structure needed for downstream computations~\cite{coreset_survey}.
The notion of geometrical coresets has been studied for neural network training~\cite{COLLIDER_LID_coreset}. However, this differs from our algorithm, as they were used to approximate the gradient of the clean data.
Here, our intention is to form a neighborhood subspace that serves as a reference for better distinguishing far-OOD samples.
Particularly, for each ID class, we select $m$ samples requiring the fewest dimensions to define their local geometry.
Empirically, LID has been linked to outlier detection; prior work~\cite{LID_adversarial,LID_OOD_correlation} shows that neural networks progressively transform data into regions of low intrinsic dimensionality, clustering clean samples together.
Therefore, the proposed algorithm aims to compute the smallest \textit{m} balls of LID to select the most centrally located ID samples.

Formaly, let $\mathcal{Z}_y $ denote the set of superpixel embeddings of the training set associated with label $y$, then the proposed coreset selection seeks to find a subset $\mathcal{Z}^{\prime}\subseteq\bigcup_{y \in \mathcal{Y}} \mathcal{Z}_y$ where:
\begin{equation}\label{eq:corset}
\mathcal{Z}^{\prime}=\bigcup_{y \in \mathcal{Y}} \mathcal{Z}_y^{\prime} \quad \text { s.t. } \quad \mathcal{Z}_y^{\prime}=\underset{T \subseteq \mathcal{Z}_y, |T|=m}{\arg \min } \sum_{\boldsymbol{z}_t \in T} \widehat{\mathrm{LID}}(\boldsymbol{\boldsymbol{z}_t}).
\end{equation}
Additionally, we obtain an associated weight set, $\mathcal{W}^{\prime} = \left\{w_t\right\}_{t=1}^{K\times m}$, where $w_t = \widehat{\mathrm{LID}}(\boldsymbol{\boldsymbol{z}_t})$ is the LID associated with the selected superpixel itself.

\subsubsection{SupLID Score}
Now we propose a novel OOD score, motivated by the fact that a data sample situated in a neighborhood lacking similar samples is likely to exhibit high intrinsic dimensionality~\cite{LID_OOD_correlation, LID_adversarial}.
While directly identifying a subspace dissimilar to OOD is challenging—due to the unknown nature of OOD data distributions—we can instead construct a subspace that is maximally dissimilar to the ID samples, leveraging the model's prior knowledge of ID data.
To achieve this, we generate a tampered set of ID samples by modulating each feature embedding in the coreset $\mathcal{Z}^{\prime}$ (given in~\Cref{eq:corset}) based on its local geometry, as defined by the corresponding weight in $\mathcal{W}^{\prime}$, such that non-salient features (i.e., those with high LID) are amplified, while salient features (low LID) are suppressed.

As a result, this low-dimensional manifold is now converted into a high-dimensional one dominated by noise, resembling a neighbourhood that shelters far-OOD-like samples. This fact  has also been observed in previous studies, where tampered samples (e.g., adversarial examples) consistently exhibit higher LID values~\cite{LID_adversarial,COLLIDER_LID_coreset}.
Accordingly,  $\mathcal{Z}^{\prime}$ forms a subspace that is effectively secluded from the ID samples.
The intention is for ID samples to exhibit relatively higher LID in this altered subspace compared to OOD samples, since a higher-dimensional neighbourhood is now required to shelter ID samples in this altered subspace.

To this end, we define the dimensionality based estimate for $D_l(\boldsymbol{x})$ as:
\begin{equation}\label{eq:dl_x}
D_l(\boldsymbol{x}) = \widehat{\mathrm{LID}^{\text{wtd}}}(\boldsymbol{z}_l)=-\left(\frac{1}{k} \sum_{i=1}^k \log \frac{r_i^{\text{wtd}}(\boldsymbol{z}_l)}{r_k^{\text{wtd}}(\boldsymbol{z}_l)}\right)^{-1},
\end{equation}
where $r_i^\text{wtd}(\boldsymbol{z}_l)$ now denotes the distance from $\boldsymbol{z}_l$ to its $i$-th nearest neighbor chosen from the weighted coreset; $\mathcal{W}^{\prime} \circ \mathcal{Z}^{\prime} =\left\{w_t \boldsymbol{z}_t\right\}_{t=1}^{K\times m}$. 
Note that the number of neighborhood queries required to estimate \Cref{eq:dl_x} is now only $K \times m \ll N \times|\mathcal{Z}|$ which we later show in~\Cref{sec:ablation} to significantly improve performance even for relatively small $m$ values on the order of $10^2$.
This is because the MLE estimator for LID typically requires on the order of 100 neighborhood samples for convergence~\cite{LID_OOD_correlation}.

Finally, we define the proposed SupLID score as $S^{SupLID}_l(\boldsymbol{x}) = S_l(\boldsymbol{x}) \cdot D_l(\boldsymbol{x})$, where $S_l(\boldsymbol{x})$ and $D_l(\boldsymbol{x})$ are given by \Cref{eq:sl_x} and \Cref{eq:dl_x}, respectively.
The per-pixel anomaly map, matching the resolution of the original image, is then obtained by broadcasting the $S^{SupLID}_l$ score to their corresponding pixels.

\begin{table*}[tb!]
\centering
\caption{Comparison of SupLID results on recent SOTA models that incorporate re-training-based approaches. The best results are bolded. Results averaged over six random runs.}
\begin{small}
\setlength{\tabcolsep}{4pt}
\renewcommand{\arraystretch}{0.9} 
\begin{tabular}{l*{5}{ccc}}
\toprule
\textbf{Method} & \multicolumn{3}{c}{\textbf{SMIYC-Anomaly}} & \multicolumn{3}{c}{\textbf{SMIYC-Obstacle}} & \multicolumn{3}{c}{\textbf{RoadAnomaly}} & \multicolumn{3}{c}{\textbf{Fishyscapes-Static}} & \multicolumn{3}{c}{\textbf{Fishyscapes-L\&F}} \\
\cmidrule(lr){2-4} \cmidrule(lr){5-7} \cmidrule(lr){8-10} \cmidrule(lr){11-13} \cmidrule(lr){14-16}
& FPR & AUP & AUR & FPR & AUP & AUR & FPR & AUP & AUR & FPR & AUP & AUR & FPR & AUP & AUR \\
\midrule
MSP~\cite{hendrycks_2017MSP} & 60.33 & 40.81 & 78.90 & 3.86 & 41.24 & 99.10 & 71.24 & 15.79 & 67.60 & 24.33 & 22.51 & 92.16 & 40.25 & 4.96 & 89.39 \\
kNN~\cite{KNN_OOD} & 69.41 & 39.42 & 77.15 & 63.75 & 0.95 & 65.35 & 82.24 & 16.98 & 65.31 & 35.50 & 8.28 & 87.31 & \textbf{21.60} & 2.47 & 92.65 \\
NNGuide~\cite{NNGuide} & 53.37 & 49.46 & 84.03 & 18.37 & 11.01 & 95.41 & 71.84 & 19.10 & 71.78 & 19.97 & 31.61 & 94.66 & 30.15 & 10.66 & \textbf{94.53} \\
Energy~\cite{EnergyOE} & 67.46 & 45.44 & 80.48 & 4.49 & 34.47 & 98.67 & 70.16 & 19.54 & 73.35 & 17.77 & 41.68 & 95.90 & 41.79 & 16.05 & 93.71 \\
\textbf{SupLID (Ours)} & \textbf{44.55} & \textbf{59.46} & \textbf{87.94} & \textbf{2.27} & \textbf{68.96} & \textbf{99.36} & \textbf{52.52} & \textbf{30.34} & \textbf{81.98} & \textbf{17.20} & \textbf{43.32} & \textbf{96.37} & 50.07 & \textbf{24.91} & 92.12 \\
\midrule
Synboost~\cite{synboost} & 30.90 & 68.80 & - & 2.80 & 81.40 & - & 59.72 & 41.83 & 85.23 & 25.59 & 66.44 & 95.87 & 31.02 & 60.58 & 96.21 \\
DenseHybrid~\cite{densehybrid} & 52.65 & 61.08 & - & 0.71 & 89.49 & - & - & - & - & 4.17 & 76.23 & 99.07 & 5.09 & 69.79 & 99.01 \\
\midrule
BE-OE~\cite{choiBalancedEnOE} & 58.95 & 45.28 & 83.68 & 11.03 & 8.31 & 95.76 & \textbf{41.46} & 41.49 & 88.31 & \textbf{1.17} & \textbf{92.49} & \textbf{99.55} & \textbf{2.94} & \textbf{67.07} & \textbf{99.03} \\
\textbf{BE-OE+SupLID (Ours)} & \textbf{39.19} & \textbf{59.14} & \textbf{89.03} & \textbf{6.06} & \textbf{60.32} & \textbf{98.01} & 36.69 & \textbf{53.54} & \textbf{91.10} & 4.01 & 84.36 & 99.04 & 8.77 & 66.94 & 98.39 \\
\midrule
MetaOOD~\cite{MetaOOD} & 17.46 & 80.76 & 95.91 & \textbf{0.43} & 94.22 & 99.90 & 10.85 & \textbf{85.19} & 97.72 & 13.98 & 73.39 & 97.56 & 37.40 & 41.36 & 93.11 \\
\textbf{MetaOOD+SupLID (Ours)} & \textbf{15.45} & \textbf{82.13} & \textbf{96.43} & 0.48 & \textbf{94.50} & 99.90 & \textbf{10.04} & 81.91 & \textbf{97.76} & \textbf{13.48} & \textbf{80.86} & \textbf{98.04} & \textbf{35.98} & \textbf{47.71} & \textbf{93.68} \\
\midrule
PEBAL~\cite{PEBAL} & 36.49 & 53.81 & 89.07 & 7.92 & 10.45 & 96.72 & 44.58 & 45.12 & 87.63 & \textbf{1.52} & \textbf{92.09} & \textbf{99.60} & \textbf{4.76} & 58.82 & \textbf{98.96} \\
\textbf{PEBAL+SupLID (Ours)} & \textbf{28.14} & \textbf{69.13} & \textbf{92.51} & \textbf{6.74} & \textbf{77.68} & \textbf{98.39} & \textbf{35.80} & \textbf{57.40} & \textbf{91.17} & 3.15 & 86.78 & 99.30 & 6.45 & \textbf{63.12} & 98.78 \\
\midrule
RPL~\cite{RPL} & 7.18 & 88.55 & 98.06 & \textbf{0.09} & \textbf{96.91} & \textbf{99.96} & 17.74 & 71.60 & 95.72 & 0.85 & 92.46 & 99.73 & \textbf{2.52} & 70.61 & \textbf{99.39} \\
\textbf{RPL+SupLID (Ours)} & \textbf{6.74} & \textbf{89.77} & \textbf{98.24} & 0.10 & 96.42 & 99.81 & \textbf{17.48} & \textbf{73.03} & \textbf{95.90} & \textbf{0.62} & \textbf{94.16} & \textbf{99.80} & 4.11 & \textbf{70.85} & 99.16 \\
\bottomrule
\end{tabular}
\end{small}
\label{tab:main_sota_retrain_results}
\end{table*}


\section{Evaluation}\label{sec:Evaluation}
This section evaluates the performance of SupLID on diverse OOD benchmarks in semantic segmentation.
We first compare SupLID with other post-hoc OOD detection scores. Then, we evaluate its performance when applied to SOTA retraining-based OOD detection models to improve their classifier-based scores.
We also conduct a thorough ablation study to analyze the components within SupLID, assess its compatibility with various classifier-based confidence scores, and examine the impact of its key hyperparameters.  
 
\subsection{Experimental Settings} \label{sec:experimental_setup}
\subsubsection{ID dataset:} Following prior work~\cite{RPL,PEBAL,MetaOOD}, we use the Cityscapes~\cite{cityscapes_dataset} dataset as the ID data.
It contains 2,975 training images of urban driving scenes at a resolution of 2048×1024 and includes 19 semantic classes.

\subsubsection{OOD datasets:} We evaluate our method on a diverse set of OOD benchmarks. The Segment-Me-If-You-Can~(SMIYC)~\cite{SMIYC_dataset} dataset includes two subsets: SMIYC-Anomaly, which contains anomalous objects in various contexts, and SMIYC-Obstacle, which contains small obstacles placed on the road. 
The public validation sets include 10 and 30 images, while the online test sets—entirely hidden from the evaluated methods—consist of 100 and 327 images for SMIYC-Anomaly and SMIYC-Obstacle, respectively.
The Fishyscapes benchmark~\cite{FS_dataset} also includes two subsets: FS-Static, consisting of 30 images with synthetic anomalous objects, and FS-Lost\&Found, containing 100 images of real road anomalies. Additionally, we evaluate on the RoadAnomaly\cite{RoadAnomaly_dataset} dataset, which includes 60 images of real-world anomalous objects.

\subsubsection{Backbone models:} As the base semantic segmentation model, we use DeepLabv3+ with a WideResNet-38 backbone, trained by NVIDIA~\cite{DeepLab_NVIDIA}, following the setup used in many recent SOTA OOD detection methods~\cite{PEBAL,RPL,MetaOOD,choiBalancedEnOE}. 

\subsubsection{Configuration:} As mentioned, SupLID requires no retraining of the original model and involves only three hyperparameters: the number of superpixels $N$, the number of neighborhood samples $k$, and the size of the ID coreset per class $m$.
We select $k = 400$ and $m = 400$ from the candidate set \{10, 20, 50, 100, 200, 300, 400, 500, 600, 800, 1000, 2000\}, and determine $N$ by dividing the total number of pixels by a factor of 200, chosen from the set $\{10, 50, 100, 150, 200, 250, 300, 350, 400, 500\}$, considering both efficiency and performance, following the validation strategy in~\cite{hendrycks_2018OE}.
The dimension of the penultimate feature layer, where LID is computed, is 304.
We further analyze the effect of hyperparameters in~\Cref{sec:hyper_para}.

\subsubsection{Evaluation metrics:} We use widely adopted evaluation metrics~\cite{PEBAL,RPL,MetaOOD} for OOD detection, including the area under the receiver operating characteristic curve (AUR), the area under the precision-recall curve (AUP), the false positive rate at 95\% true positive rate (FPR), and the F1 score.
Our method does not compromise the ID segmentation performance, achieving the same mean Intersection over Union (mIoU) as the original model.

\begin{figure}[tb!]
  \centering
  \begin{subfigure}[t]{0.495\linewidth}
    \centering
    \includegraphics[width=\linewidth]{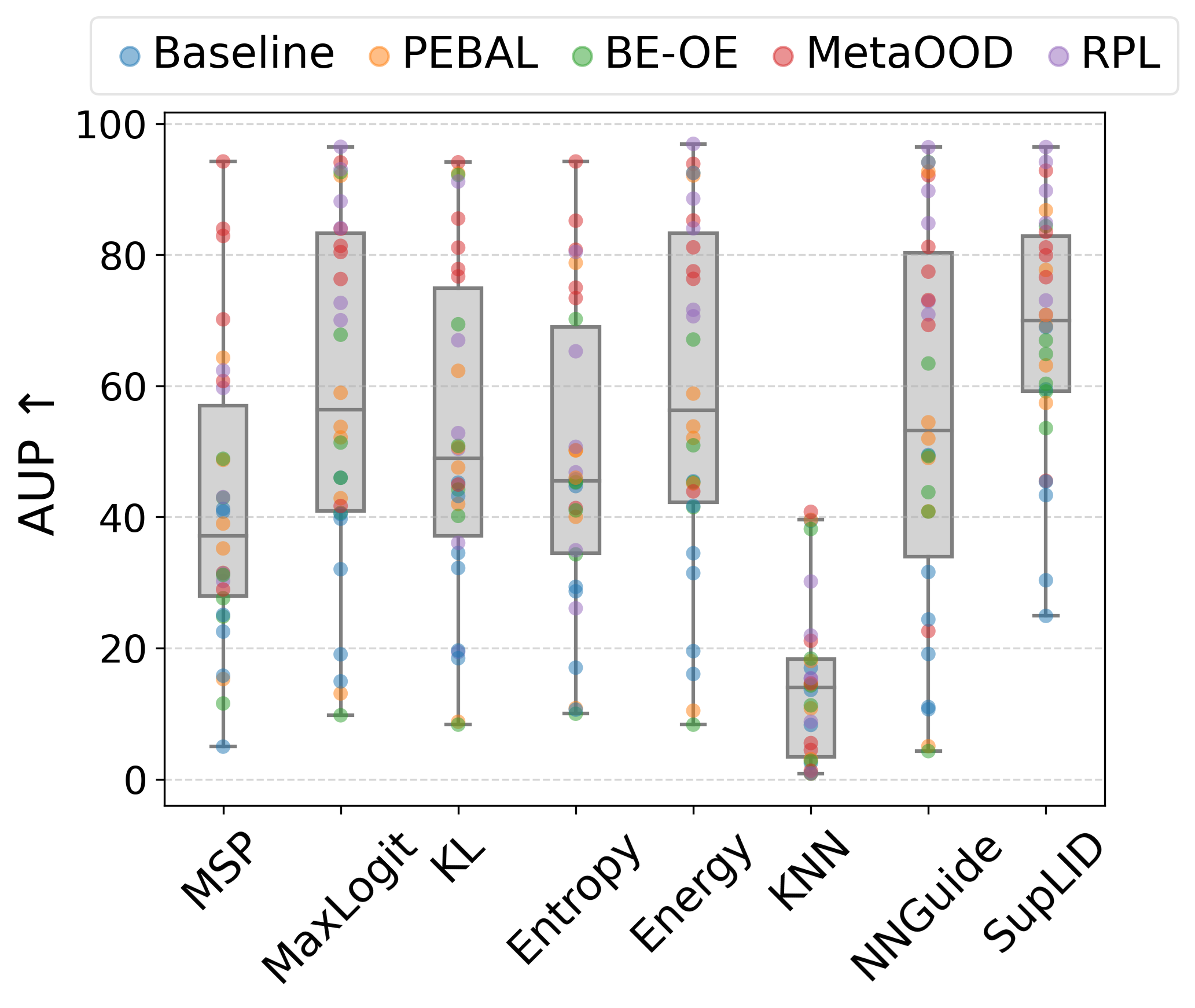}
    \label{fig:aup_plot}
  \end{subfigure}
  \begin{subfigure}[t]{0.495\linewidth}
    \centering
    \includegraphics[width=\linewidth]{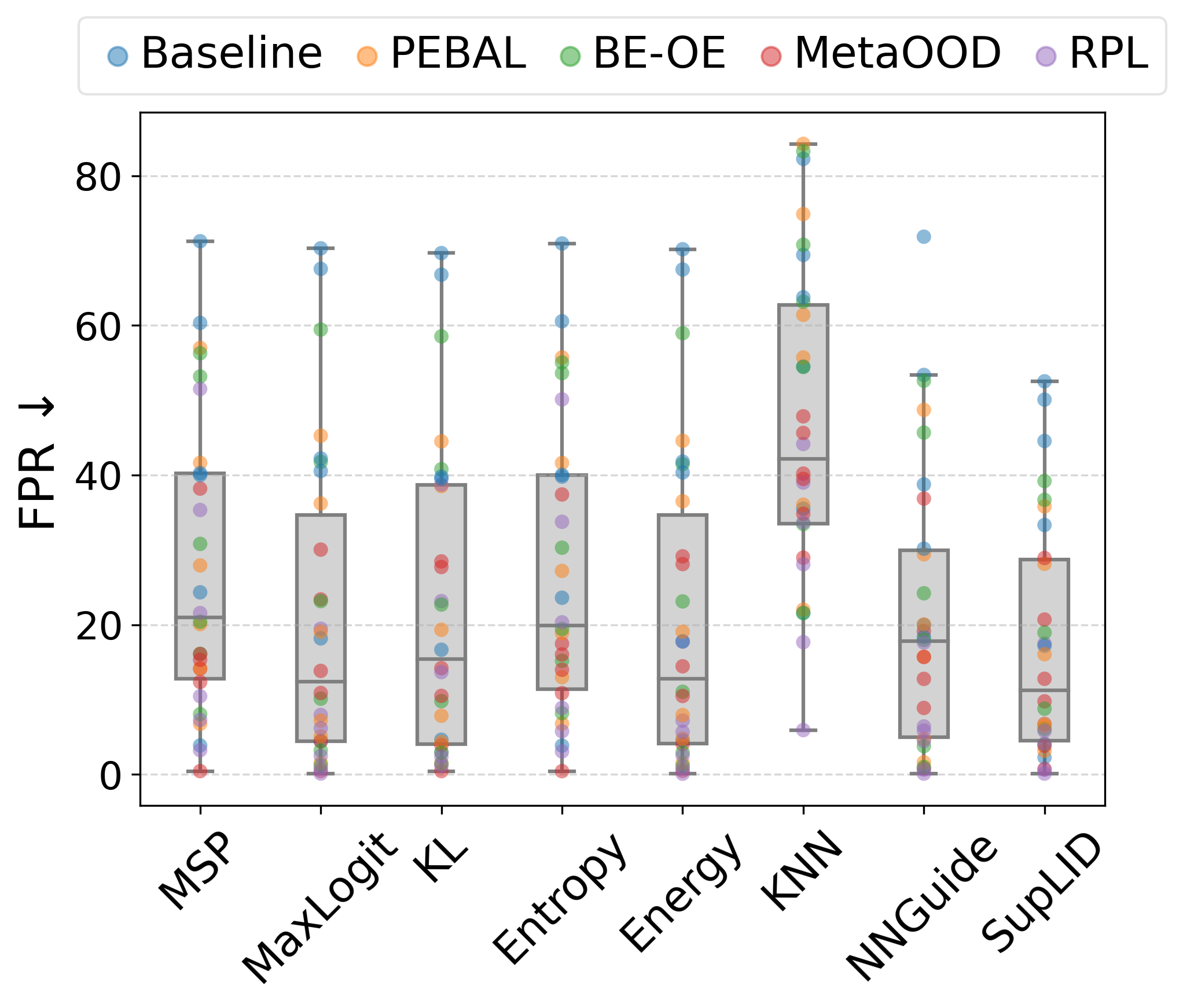}
    \label{fig:fpr_plot}
  \end{subfigure}
   \vspace{-0.15in}
  \caption{Summary of SupLID performance and other OOD scores on five OOD benchmarks across different models.}
  \label{fig:summery_performance}
   \vspace{-0.1in}
\end{figure}

\subsection{Main Results} \label{sec:main_results}

\subsubsection{Comparison with post-hoc detection scores} \label{sec:comparison_post_hoc}
We first compare SupLID with SOTA post-hoc OOD detection scores applied to the base semantic segmentation model. Specifically, we evaluate against classifier-based confidence scores, including MSP~\cite{hendrycks_2017MSP}, MaxLogit~\cite{MaxLogit}, KL~\cite{MaxLogit}, entropy~\cite{MetaOOD}, and energy~\cite{EnergyOE}. Additionally, we compare SupLID with non-parametric methods such as kNN~\cite{KNN_OOD}, Mahalanobis~\cite{lee_2018_mahalanobis}, and NNGuide~\cite{NNGuide}, which leverage feature-space distance information. To obtain pixel-wise OOD scores from the latter group, we apply the same coreset selection and superpixel-based aggregation steps used in our proposed method.
Results in \Cref{tab:main_posthoc_results} indicate that SupLID achieves a new SOTA performance by significantly outperforming other post-hoc scores across all evaluation metrics.
Notably, SupLID surpasses NNGuide, which relies on kNN to guide the classifier score but does not incorporate knowledge of the data's intrinsic dimensionality.

\begin{figure*}[tb!]
    \centering
    \setlength{\tabcolsep}{1pt}
   \begin{tabular}{@{}ccccc@{}}
        \begin{subfigure}{0.2\textwidth}
            \includegraphics[width=\linewidth]{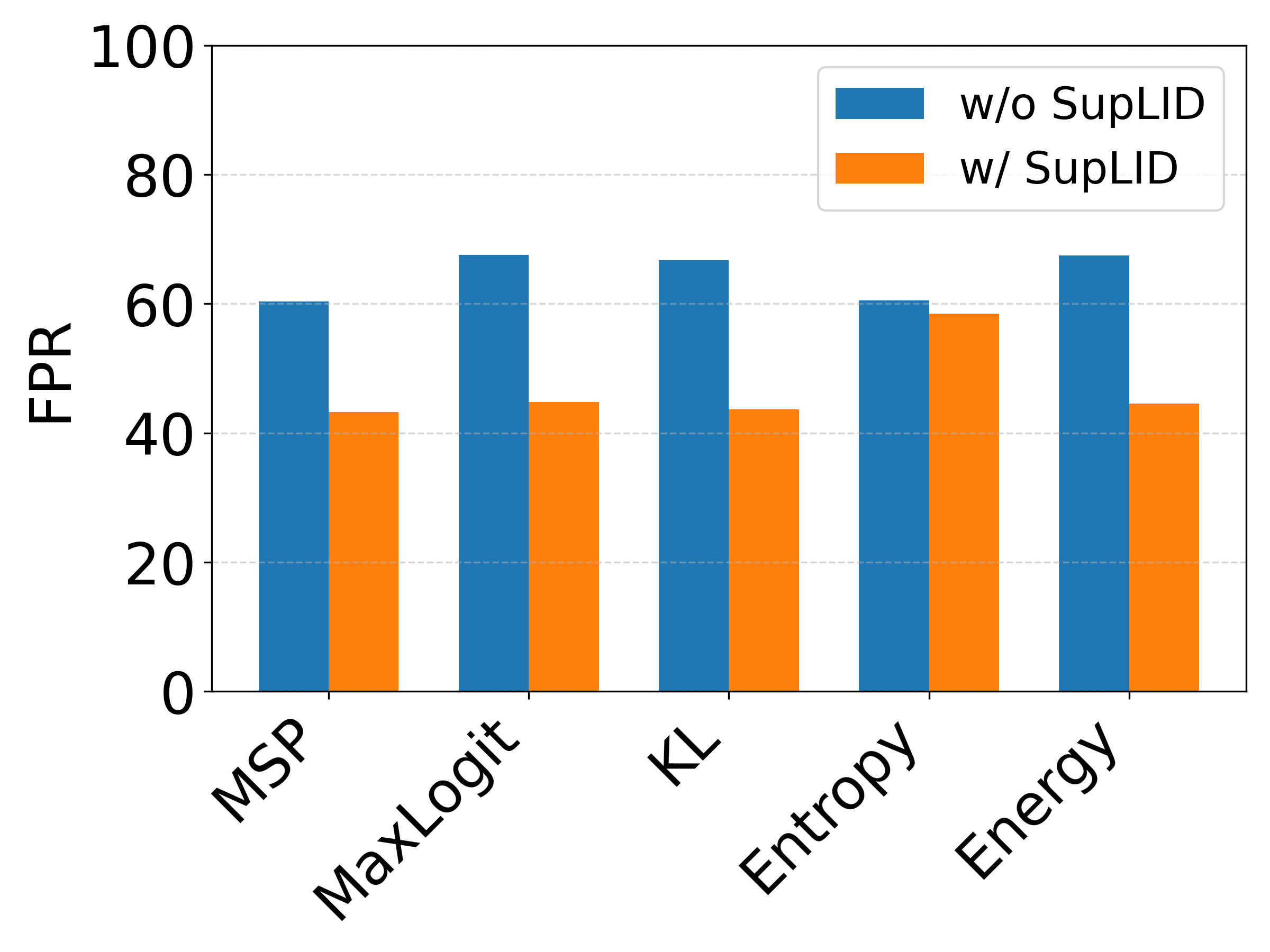}
            \caption{Baseline (DeepLabv3+)}
        \end{subfigure} &
        \begin{subfigure}{0.2\textwidth}
            \includegraphics[width=\linewidth]{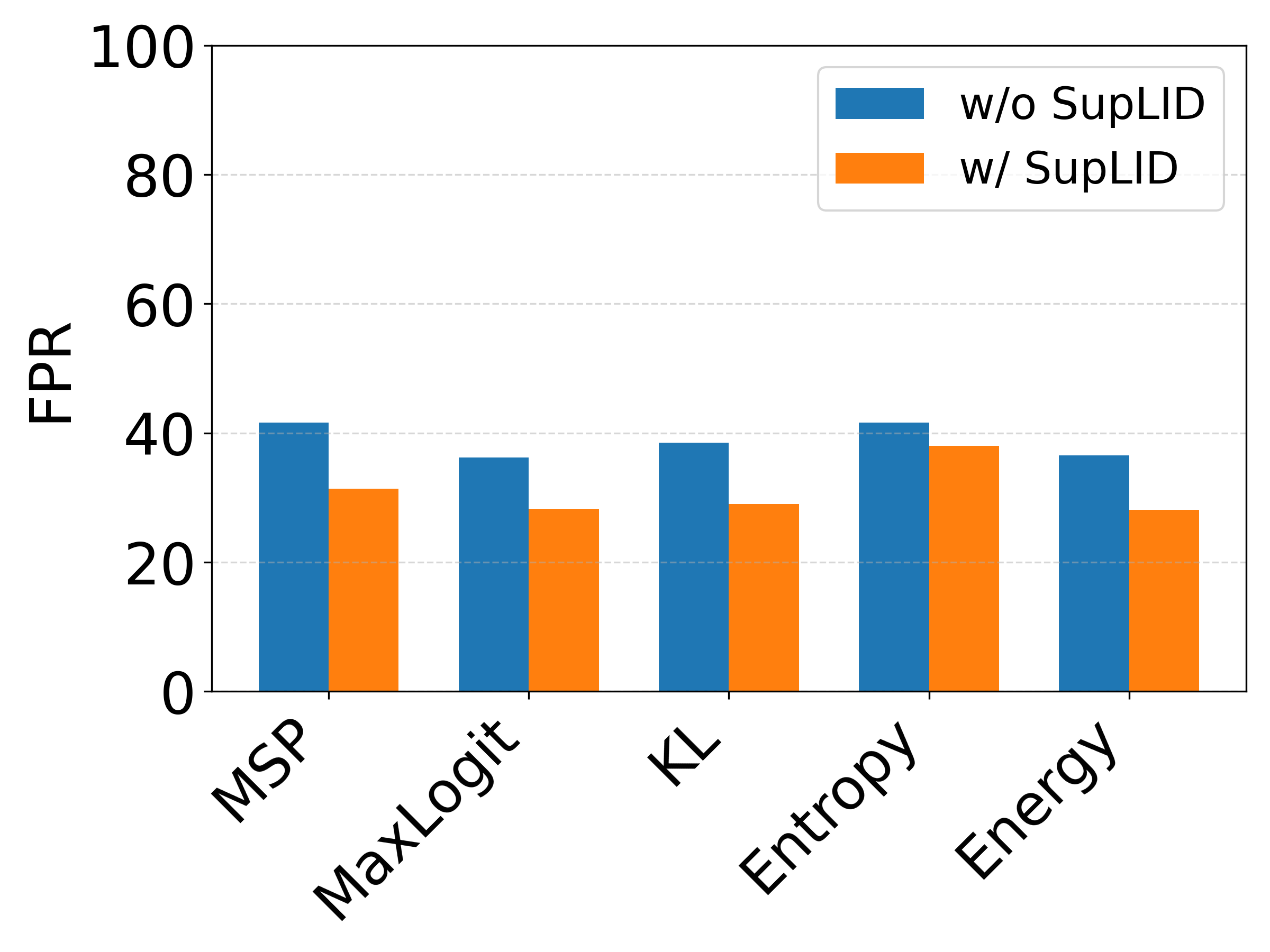}
            \caption{PEBAL~\cite{PEBAL}}
        \end{subfigure} &
        \begin{subfigure}{0.2\textwidth}
            \includegraphics[width=\linewidth]{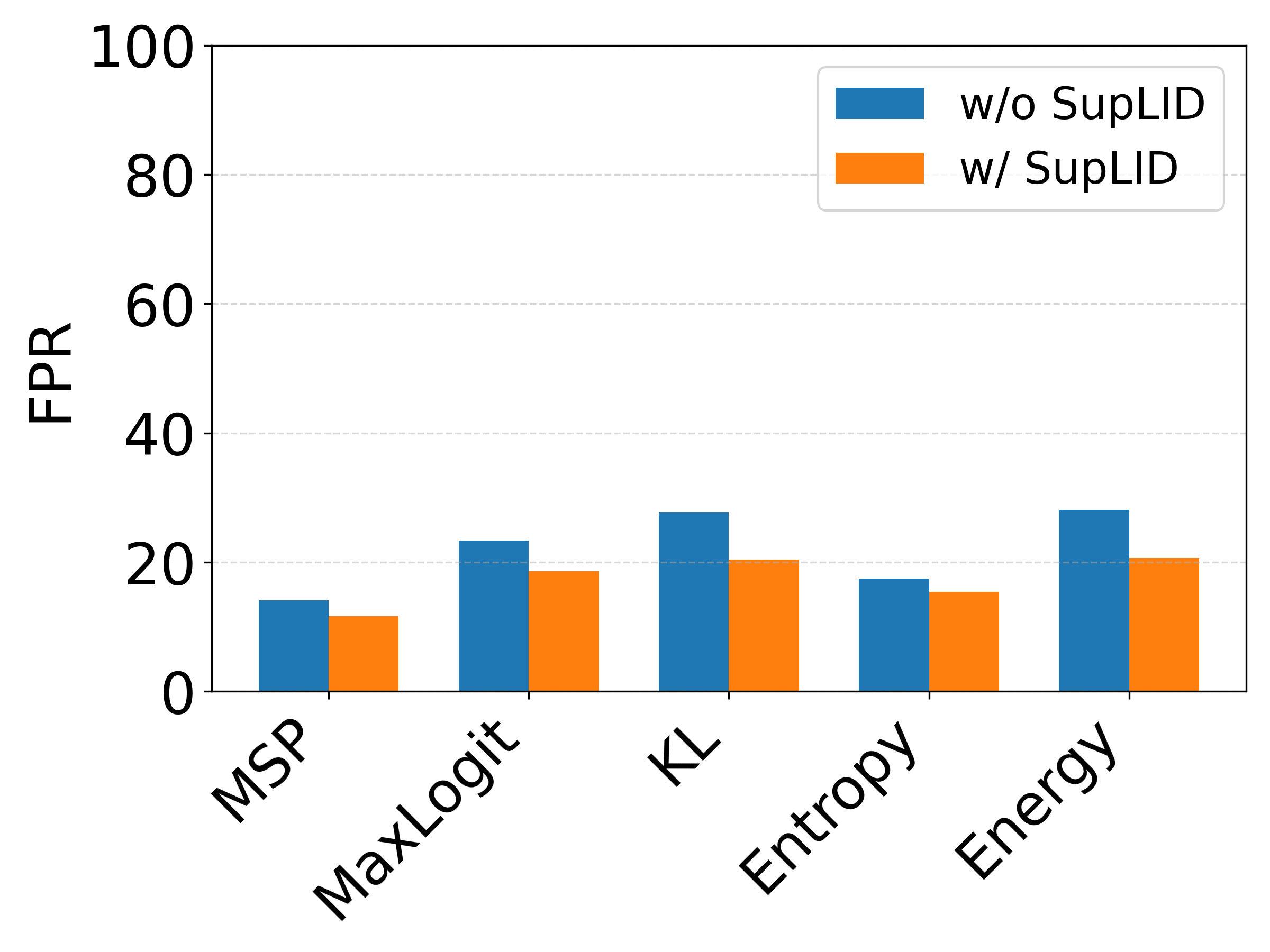}
            \caption{MetaOOD~\cite{MetaOOD}}
        \end{subfigure} &
        \begin{subfigure}{0.2\textwidth}
            \includegraphics[width=\linewidth]{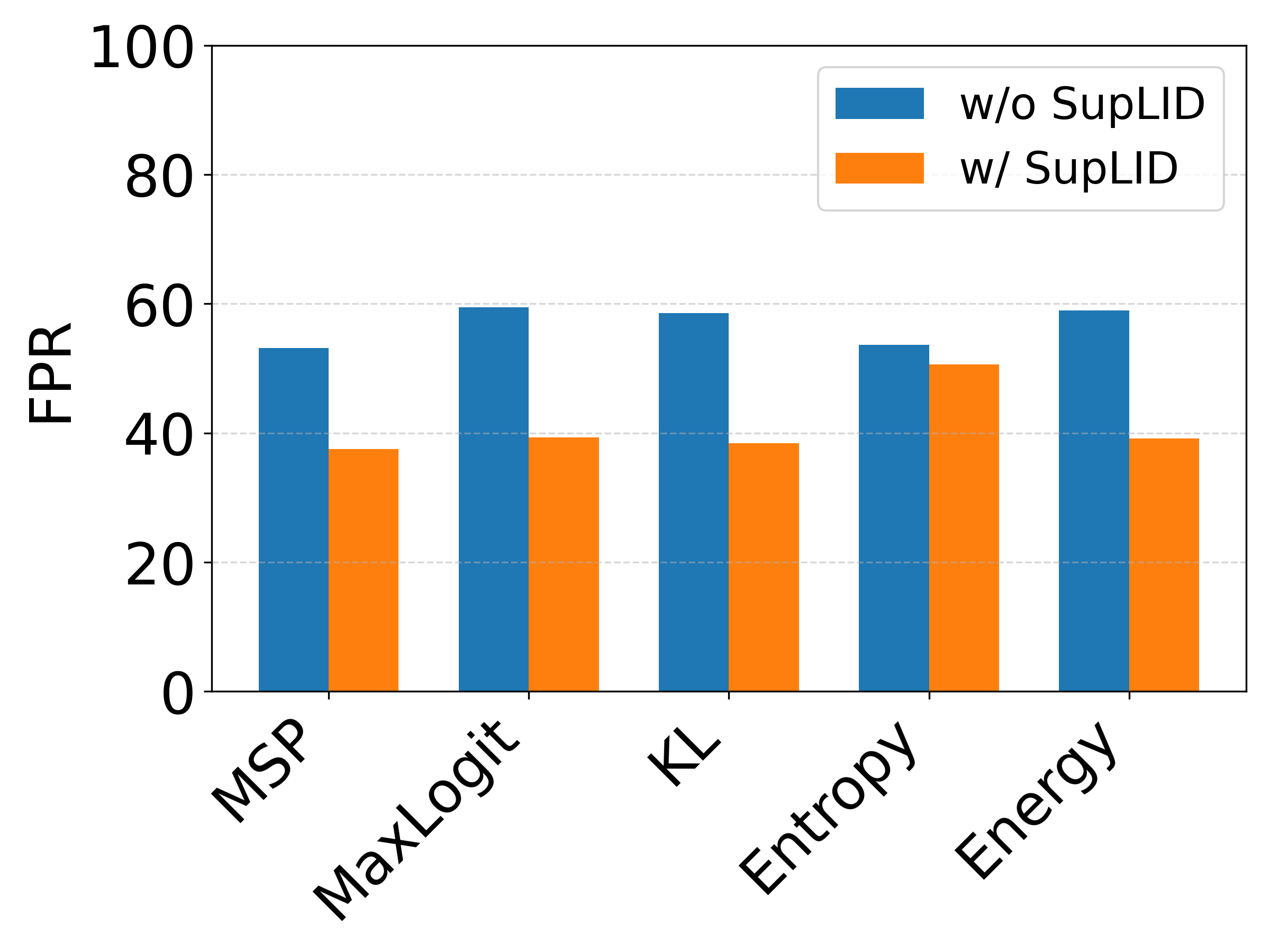}
            \caption{BE-OE~\cite{choiBalancedEnOE}}
        \end{subfigure} &
        \begin{subfigure}{0.2\textwidth}
            \includegraphics[width=\linewidth]{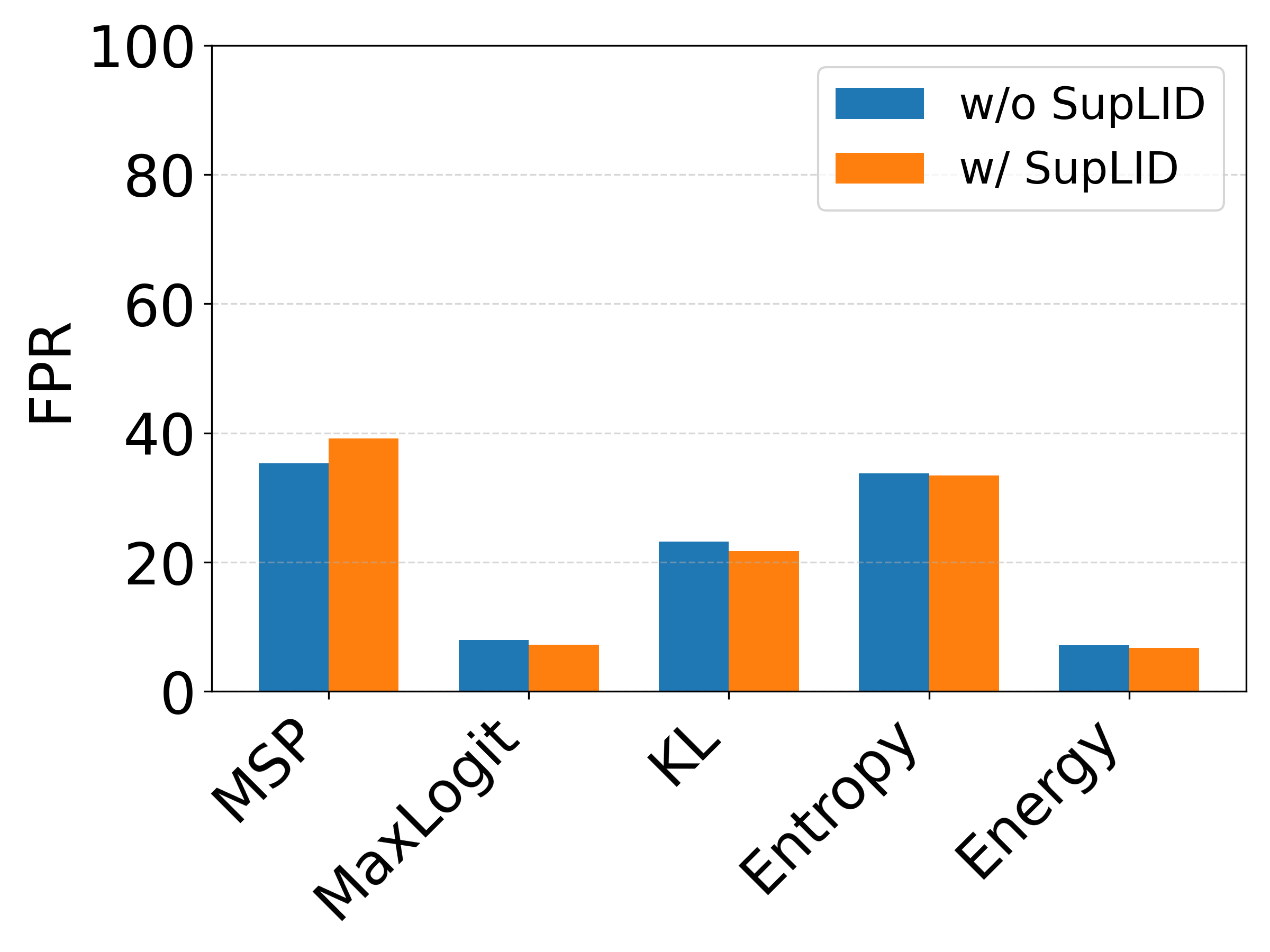}
            \caption{RPL~\cite{RPL}}
        \end{subfigure} \\
    \end{tabular}
     \vspace{-0.1in}
    \caption{Compatibility of SupLID with various classifier-based scores across different baselines on SMIYC-Anomaly dataset.}
    \label{fig:bar_plot_smiyc}
\end{figure*}

\subsubsection{SOTA performance with re-training methods} \label{sec:on_sota}
We next evaluate SupLID on recent SOTA models that incorporate OOD re-training, including PEBAL~\cite{PEBAL}, MetaOOD~\cite{MetaOOD}, Balanced EnergyOE (BE-OE)~\cite{choiBalancedEnOE}, and RPL~\cite{RPL} as shown in~\Cref{tab:main_sota_retrain_results}. For each method, we use its default confidence score to compute the SupLID score.
We also include other recent OOD detection methods (SynBoost~\cite{synboost}, DenseHybrid~\cite{densehybrid}) based on reconstruction or hybrid approaches in the same table for comparison (values taken from their respective papers; ‘–’ indicates results not reported).

As seen, SupLID consistently improves OOD detection performance across all evaluated baselines.
The improvement is particularly significant on the SMIYC benchmark, which features new contexts outside the training ID set, demonstrating SupLID’s robustness to context shifts.
The improvement is less substantial for RPL, likely because its performance is already saturated due to its use of context-aware contrastive learning. Nevertheless, RPL requires complex retraining, whereas SupLID offers a plug-and-play solution that can be directly applied to even lower-performing models to boost OOD detection performance, without any additional retraining or architectural modifications.

To further demonstrate the efficacy of SupLID, we evaluate it on the official black-box test sets\footnote{\label{ft:smiyc}https://segmentmeifyoucan.com/leaderboard}, as shown in~\Cref{tab:smiyc_official}.
As observed, our method improves the baseline metrics by a significant margin.
We also include a summary of the performance evaluation across all these models on five different OOD datasets in~\Cref{fig:summery_performance}. 
SupLID shows more stable performance and achieves a better average score than all other detection methods in the figure.

\begin{table}[tb!]
\centering
\caption{Comparison of SupLID results on SMIYC official test benchmark. The best results are bolded.}
\begin{small}
\setlength{\tabcolsep}{3.5pt}
\begin{tabular}{lccc|ccc}
\toprule
\textbf{Method} & \multicolumn{3}{c|}{\textbf{SMIYC-Anomaly}} & \multicolumn{3}{c}{\textbf{SMIYC-Obstacle}} \\
\cmidrule(lr){2-4} \cmidrule(lr){5-7}
& FPR & AUP & F1 & FPR & AUP & F1 \\
\midrule
Energy~\cite{EnergyOE} & 72.96 & 33.66 & 6.56 & 15.54 & 24.81 & 11.80 \\
\textbf{SupLID (Ours)} & \textbf{35.48} & \textbf{55.84} & \textbf{16.28} & \textbf{11.72} & \textbf{59.77} & \textbf{33.04} \\
\midrule
PEBAL~\cite{PEBAL} & 40.73 & 49.33 & 14.58 & 12.69 & 4.95 & 5.72 \\
\textbf{PEBAL+SupLID (Ours)} & \textbf{27.81} & \textbf{66.72} & \textbf{24.48} & \textbf{11.47} & \textbf{56.78} & \textbf{39.96} \\
\midrule
MetaOOD~\cite{MetaOOD} & 14.96 & 85.46 & 34.56 & 0.75 & \textbf{84.88} & 52.47 \\
\textbf{MetaOOD+SupLID (Ours)} & \textbf{10.26} & \textbf{85.65} & \textbf{37.05} & 0.81 & 82.95 & \textbf{56.27} \\
\midrule
RPL~\cite{RPL} & 11.66 & 83.45 & 30.09 & \textbf{0.58} & 86.00 & 56.61 \\
\textbf{RPL+SupLID (Ours)} & \textbf{10.39} & \textbf{84.93} & \textbf{36.54} & 0.62 & \textbf{86.32} & \textbf{58.14} \\
\bottomrule
\end{tabular}
\end{small}
\label{tab:smiyc_official}
\vspace{-0.05in}
\end{table}

\subsection{Ablation} \label{sec:ablation}
We conduct an ablation study with three main objectives: (1) To evaluate the compatibility of SupLID with various classifier-based confidence scores, beyond the model's default scores used in our main experiments; (2) to analyze the contribution of each individual component within the SupLID; and (3) to analyze the sensitivity to its key hyperparameters: $k$, $m$ and $N$.  

\subsubsection{Compatibility of SupLID with various classifier-based confidence scores} \label{sec:ablation_on_posthoc}
In \Cref{tab:main_sota_retrain_results}, we apply SupLID on each baseline using its originally proposed confidence score. Additionally, we conduct a thorough analysis using other commonly used classifier-based confidence scores, including MSP, MaxLogit, KL, energy, and entropy, across all baselines.
Results in~\Cref{fig:bar_plot_smiyc} indicate that SupLID consistently improves the performance of all these confidence scores.
Detailed results for all baselines and OOD datasets are provided in~\Cref{tab:ablation_on_other_scores} in the Appendix.
Furthermore, it consistently outperforms both kNN and NNGuide across all cases, as evidenced by the earlier results in~\Cref{tab:main_posthoc_results} .

\begin{table*}[tb!]
\centering
\caption{Ablation study on SupLID components on five different OOD datasets. Results averaged over six random runs.}
\setlength{\tabcolsep}{3pt}
\begin{small}
\begin{tabular}{l*{5}{ccc}|ccc}
\toprule
\textbf{} 
& \multicolumn{3}{c}{\textbf{SMIYC-Anomaly}} 
& \multicolumn{3}{c}{\textbf{SMIYC-Obstacle}} 
& \multicolumn{3}{c}{\textbf{RoadAnomaly}} 
& \multicolumn{3}{c}{\textbf{FS-Static}} 
& \multicolumn{3}{c}{\textbf{FS-L\&F}} 
& \multicolumn{3}{c}{\textbf{Average}} \\
\cmidrule(lr){2-4} \cmidrule(lr){5-7} \cmidrule(lr){8-10} \cmidrule(lr){11-13} \cmidrule(lr){14-16} \cmidrule(lr){17-19}
& FPR & AUP & AUR 
& FPR & AUP & AUR 
& FPR & AUP & AUR 
& FPR & AUP & AUR 
& FPR & AUP & AUR 
& FPR & AUP & AUR \\
\midrule
Energy alone & 67.46 & 45.44 & 80.48 & 4.49 & 34.47 & 98.67 & 70.16 & 19.54 & 73.35 & 17.77 & 41.68 & 95.90 & 41.79 & 16.05 & 93.71 & 40.34 & 31.44 & 88.43 \\
Energy w/ superpixel aggr. & 66.74 & 46.35 & 80.89 & 4.77 & 33.36 & 98.63 & 68.14 & 19.69 & 73.88 & 15.78 & 53.72 & 96.59 & 45.16 & 16.61 & 93.56 & 40.12 & 33.94 & 88.71 \\
\midrule
LID alone & 64.16 & 39.64 & 79.03 & 44.92 & 4.09 & 83.80 & 78.60 & 15.73 & 65.32 & 47.19 & 11.53 & 85.18 & 36.09 & 5.11 & 90.95 & 54.19 & 15.22 & 80.86 \\
LID w/ scaling of the coreset & 64.58 & 54.21 & 81.95 & 26.18 & 74.31 & 93.86 & 68.92 & 40.33 & 78.53 & 49.30 & 12.03 & 83.30 & 42.00 & 6.52 & 89.89 & 50.20 & 37.48 & 85.51 \\
\midrule
w/o LID & 62.63 & 49.46 & 82.54 & 2.81 & 53.25 & 99.16 & 61.84 & 22.45 & 76.98 & 15.10 & 51.13 & 96.70 & 51.90 & 22.19 & 92.99 & 38.86 & 39.70 & 89.68 \\
w/o coreset - random ID set & 54.71 & 54.92 & 85.39 & 2.44 & 63.65 & 99.32 & 55.66 & 26.56 & 79.85 & 12.97 & 46.94 & 96.96 & 51.50 & 22.86 & 92.39 & 35.46 & 42.99 & 90.78 \\
w/o coreset - energy ID set & 53.95 & 53.94 & 85.17 & 2.47 & 65.42 & 99.34 & 55.61 & 27.13 & 80.14 & 12.76 & 47.31 & 97.03 & 53.35 & 22.17 & 92.13 & 35.63 & 43.19 & 90.76 \\
w/o coreset - diverse ID set & 54.01 & 54.89 & 85.50 & 2.50 & 62.71 & 99.31 & 56.23 & 26.04 & 79.44 & 13.77 & 46.27 & 96.83 & 51.07 & 22.56 & 92.46 & 35.52 & 42.50 & 90.71 \\
\midrule
\textbf{SupLID} & 44.55 & 59.46 & 87.94 & 2.27 & 68.96 & 99.36 & 52.52 & 30.34 & 81.98 & 17.20 & 43.32 & 96.37 & 50.07 & 24.91 & 92.12 & \textbf{33.32} & \textbf{45.40} & \textbf{91.55} \\
\bottomrule
\end{tabular}
\end{small}
\label{tab:component_ablation}
\end{table*}

\subsubsection{Importance of the components of SupLID} \label{sec:ablation_components} 
\begin{figure}
    \centering
    \setlength{\tabcolsep}{0pt}
    \renewcommand{\arraystretch}{0} 
    \begin{tabular}{ccccc}
            \includegraphics[width=0.33\linewidth]{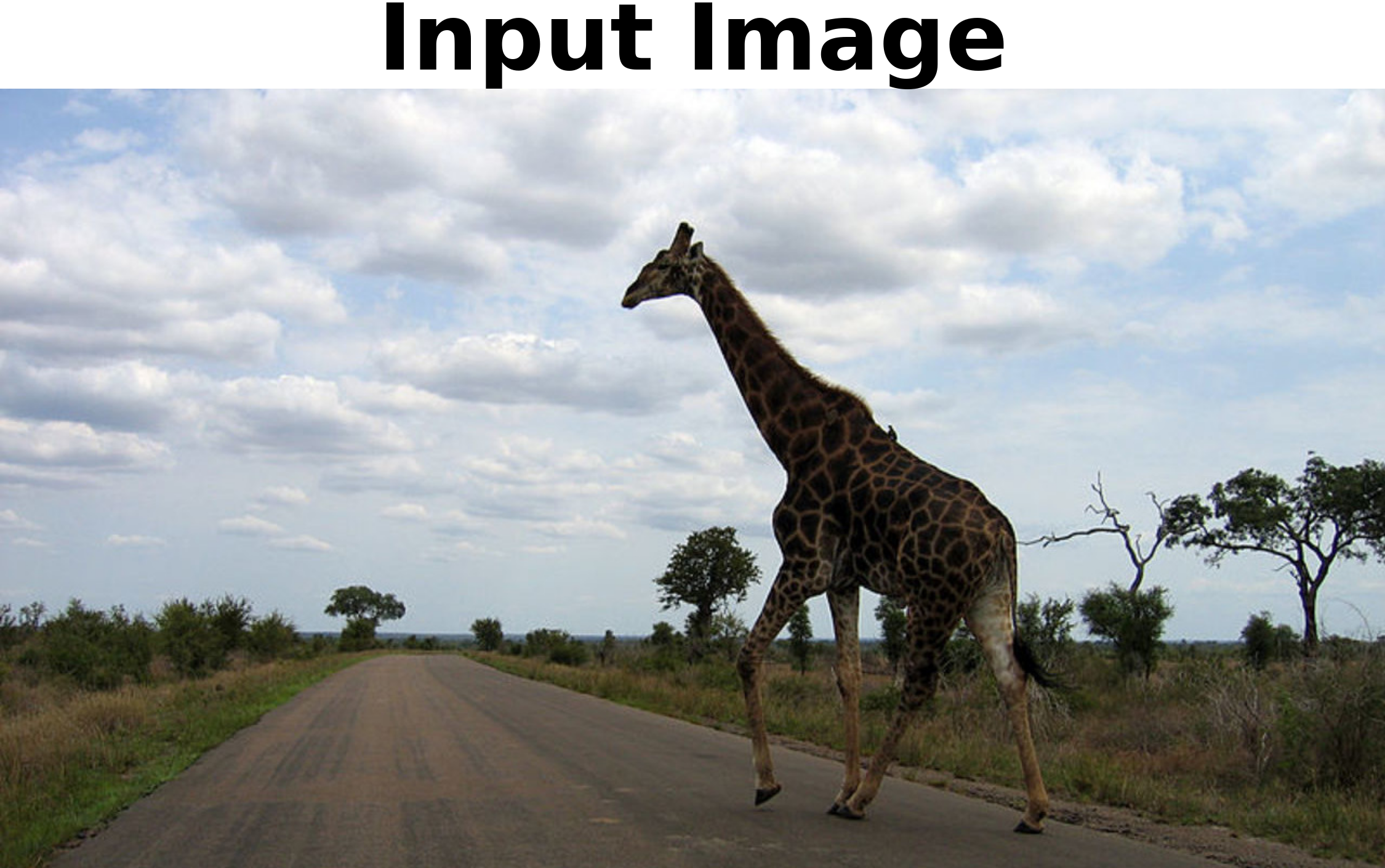} &
        \includegraphics[width=0.33\linewidth]{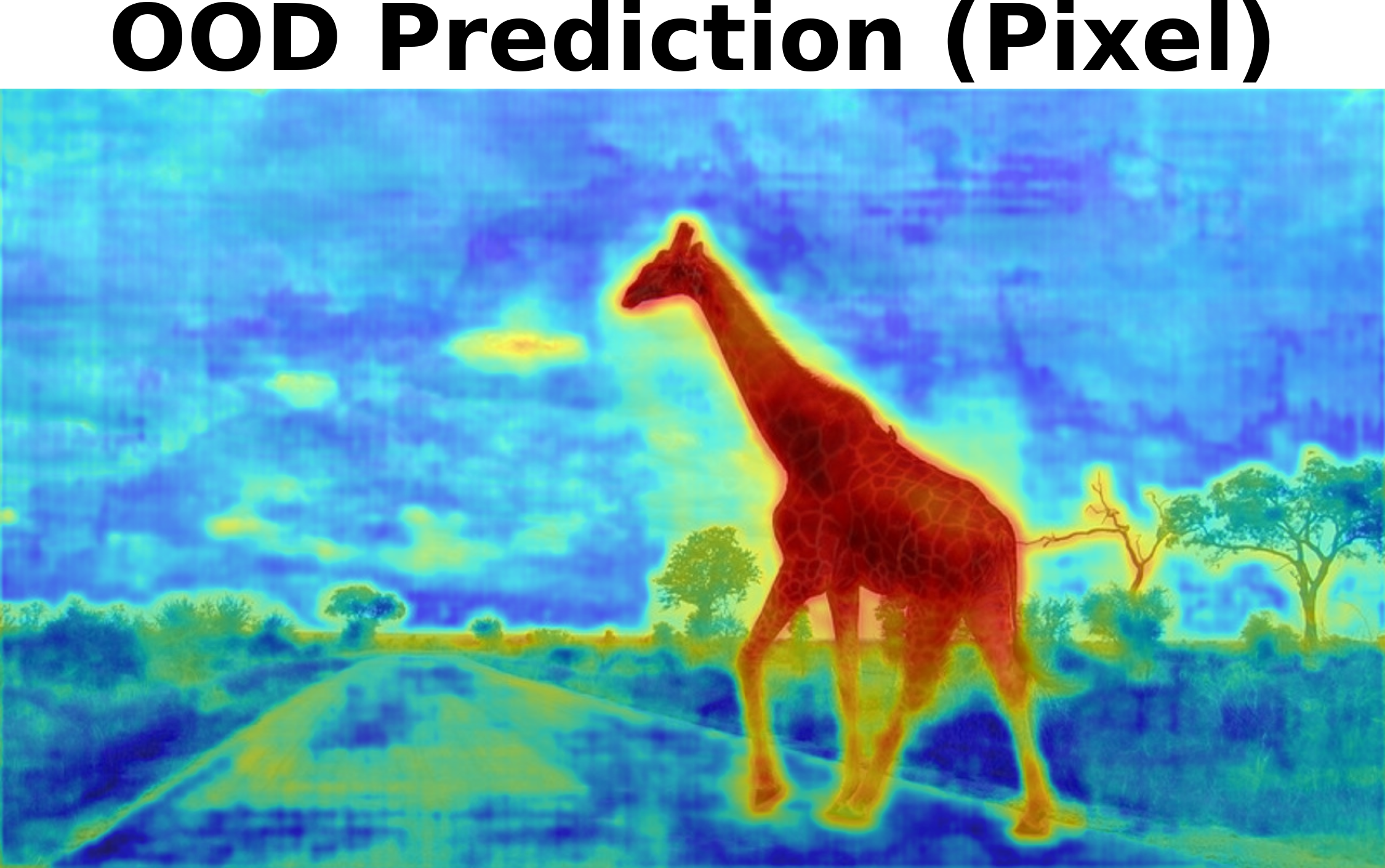} &
        \includegraphics[width=0.33\linewidth]{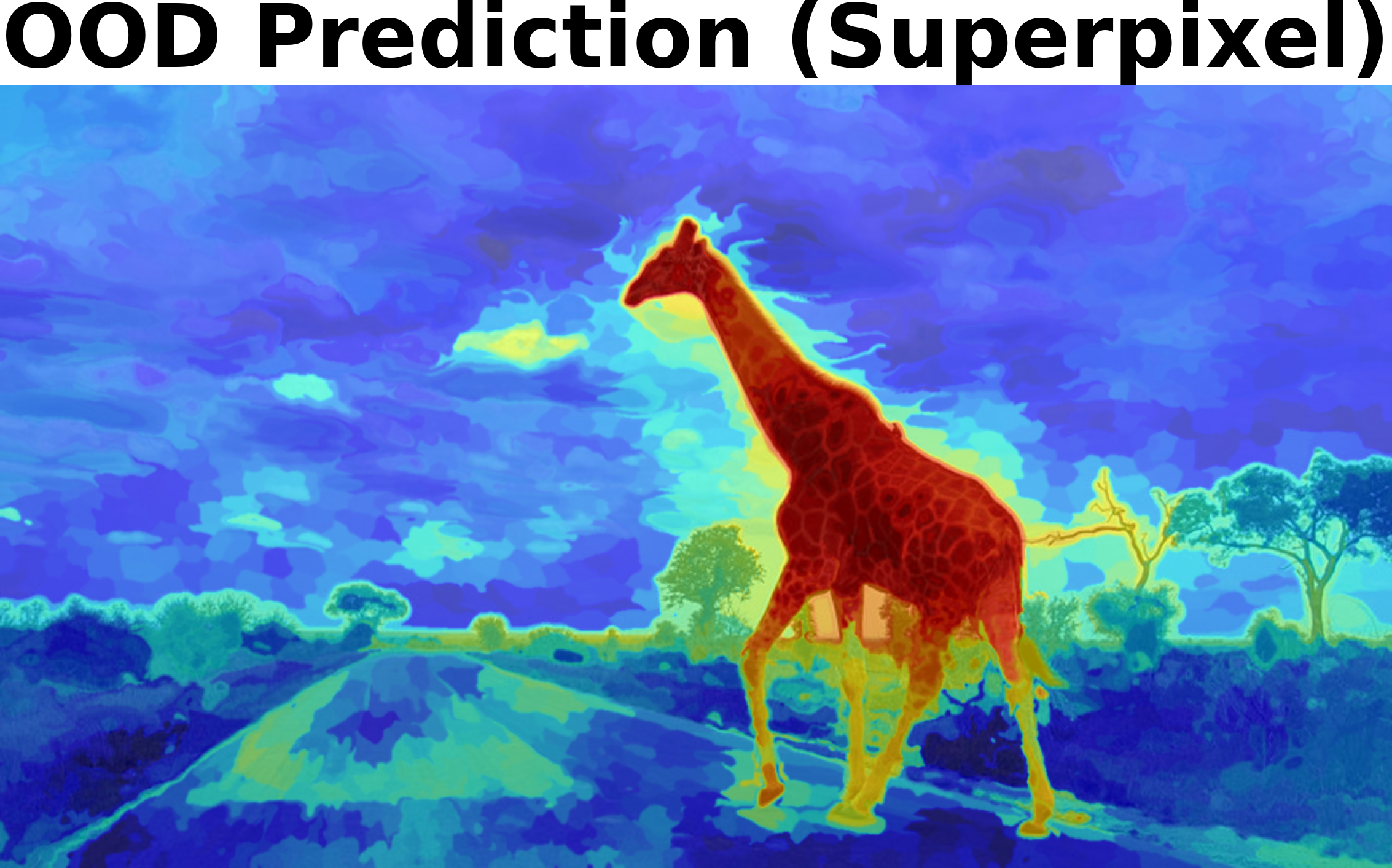} \\
        \includegraphics[width=0.33\linewidth]{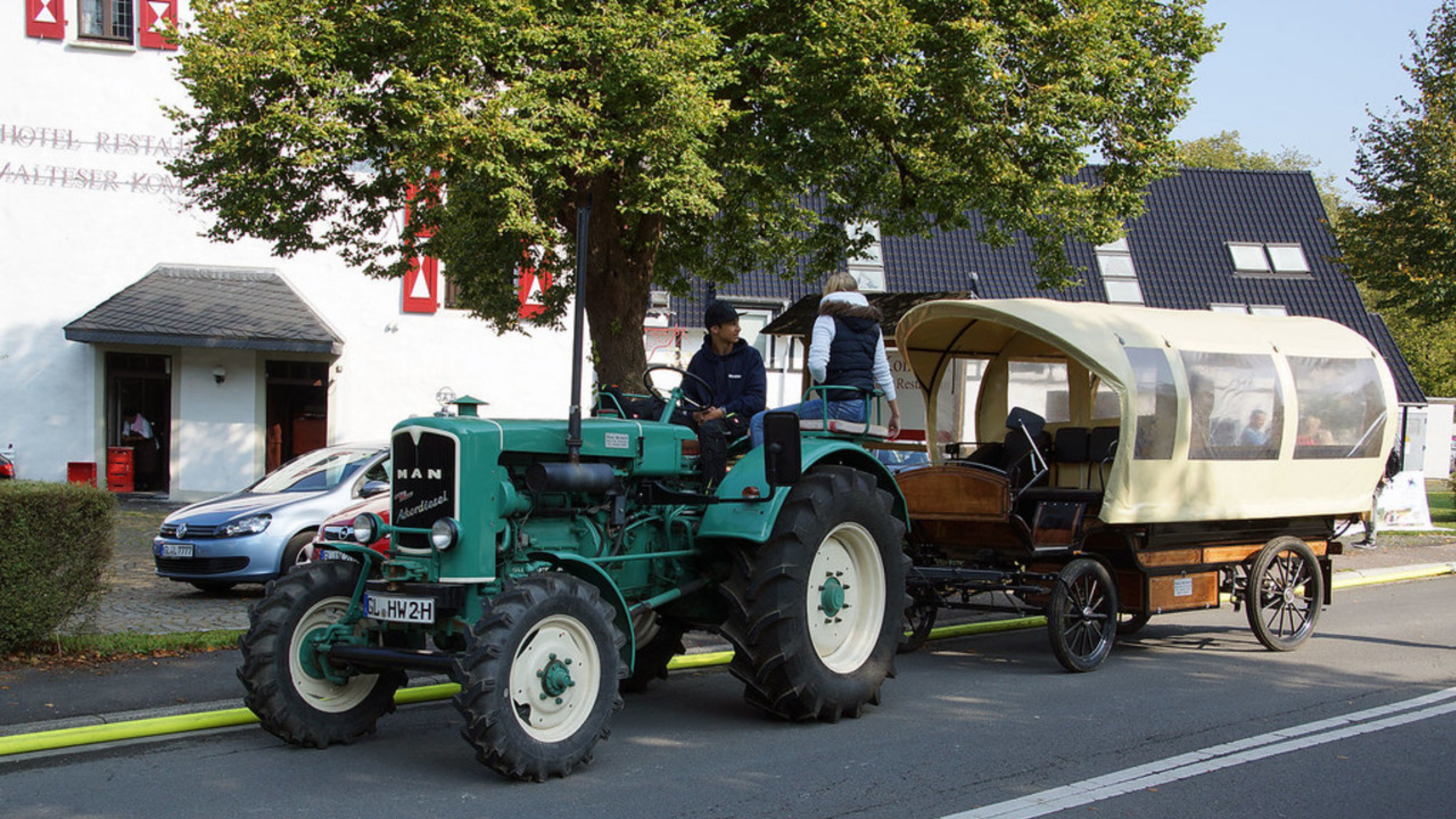} &
        \includegraphics[width=0.33\linewidth]{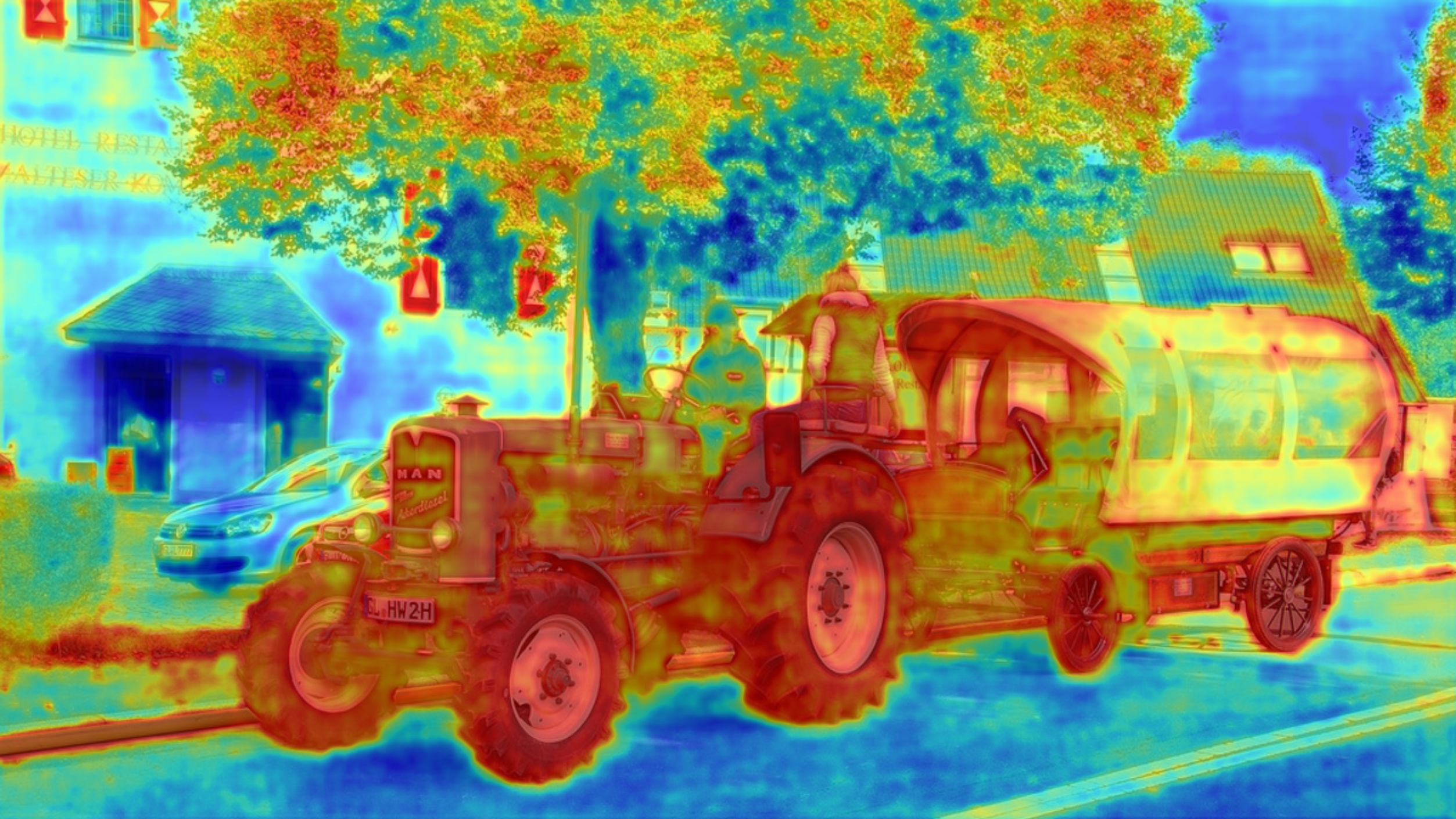} &
        \includegraphics[width=0.33\linewidth]{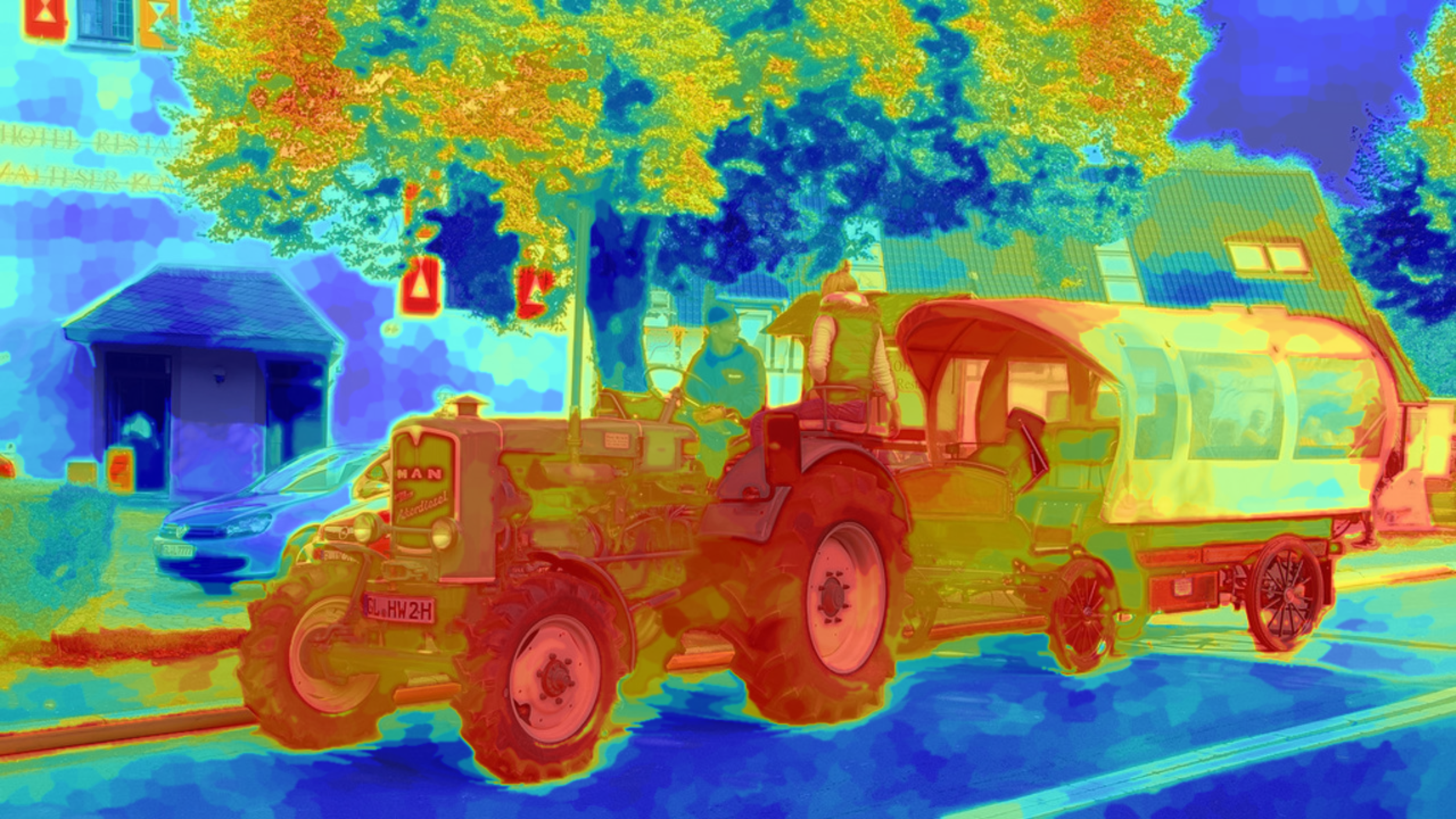} &
    \end{tabular}
   \caption{OOD prediction maps at pixel and superpixel levels based on energy~\cite{EnergyOE} of the baseline. Superpixel-level aggregation produces sharper anomaly boundaries and improves the consistent classification of neighboring object pixels.}
    \label{fig:sup_masks}
\end{figure}

SupLID enhances classifier-based confidence scores by incorporating additional geometrical information (LID) from the feature space of a semantic segmentation model.
To achieve this, SupLID constructs a geometrical coreset from the training ID data and aggregates scores based on superpixels.
In \Cref{tab:component_ablation}, we evaluate the contribution of each of these components.
First, we observe that superpixel-based aggregation not only improves computational efficiency but also boosts detection performance (`Energy alone' vs. `Energy w/ superpixel aggr.').
This can be further evidenced by the qualitative examples given in~\Cref{fig:sup_masks}.

Next, we find that the LID value alone is not a strong OOD detection score, probably due to its inherently local nature, even after applying the scaling step on the original coreset (`LID w/ scaling of the coreset').
However, the scaling still improves performance compared to using the unscaled coreset, indicating its effectiveness in refining feature-space representations. 

We then assess the importance of the geometrical coreset for SupLID by first replacing it with a randomly selected ID set of the same size (`w/o coreset - random ID set').
Additionally, we explore two alternative selection strategies: (1) constructing the coreset using the $k$ ID samples with the lowest energy values from each class, aiming to capture the most salient ID samples; and (2) selecting the coreset via diversity maximization~\cite{diverse_coreset}, where $k$ cluster centers are chosen per ID class to form a diverse set.
As shown, both approaches (`w/o coreset - energy ID set' and `w/o coreset - diverse ID set') improve performance over the base confidence score. However, their results remain suboptimal compared to the geometrical coreset used in SupLID.
Finally, we replace the LID term for $D_l(\boldsymbol{x})$ in~\Cref{eq:dl_x} with the average of the minimum $k$-nearest neighbor Euclidean distances (`w/o LID').
This yields only a marginal improvement over the base confidence score, highlighting the importance of intrinsic dimensionality estimation in SupLID and demonstrating its ability to effectively combine the complementary strengths of classifier-based scores and geometrical structure in the feature space.

\begin{figure}[tb!]
\vspace{0.1in}
  \centering
  \includegraphics[width=0.7\linewidth]{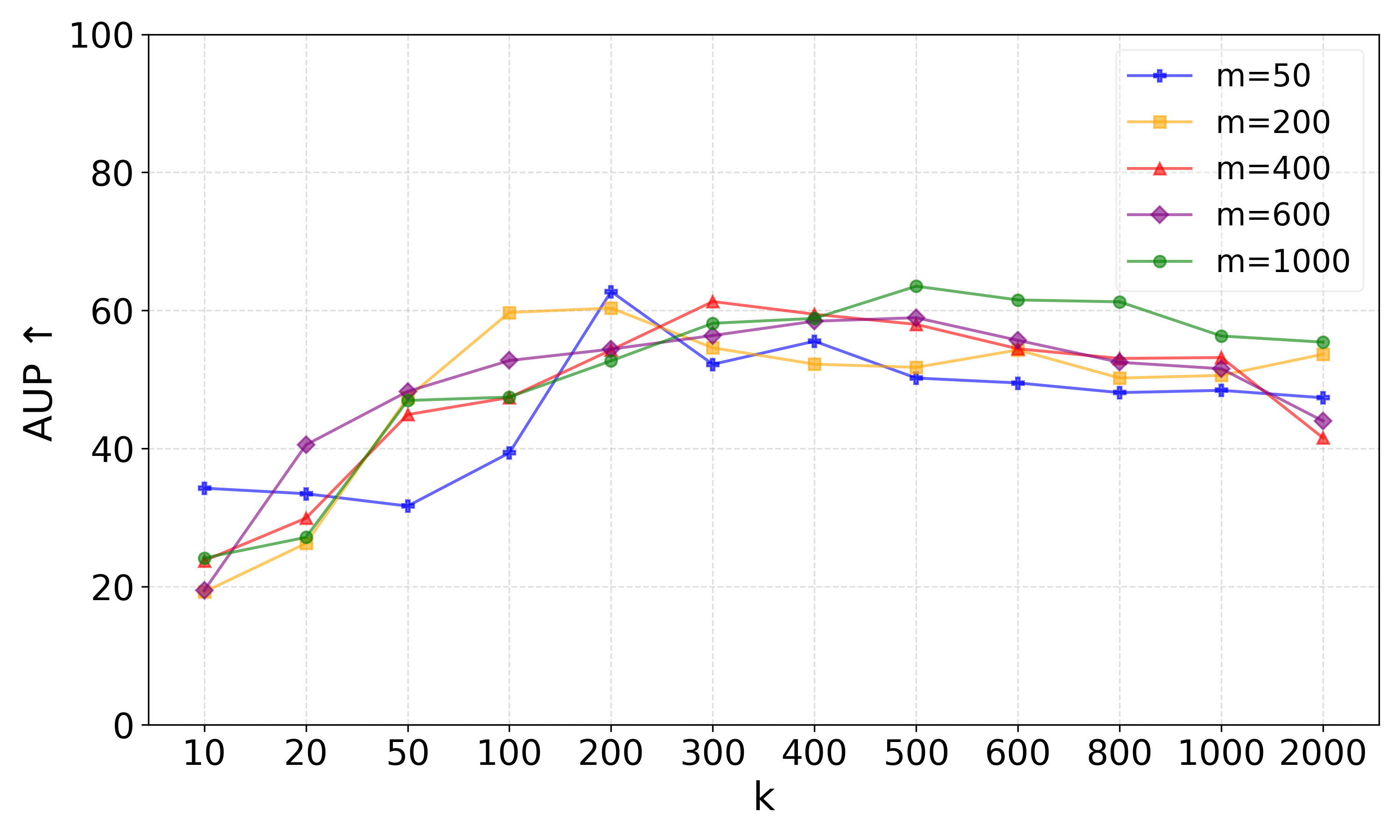}
     \vspace{-0.05in}
\caption{Variation of OOD detection performance across different values of $k$ for various coreset sizes $m$. Averaged over six random runs on SMIYC-Anomaly.}
  \label{fig:k_analysis}
   \vspace{0.05in}
\end{figure}

\begin{figure}[tb!]
  \centering
  \includegraphics[width=0.7\linewidth]{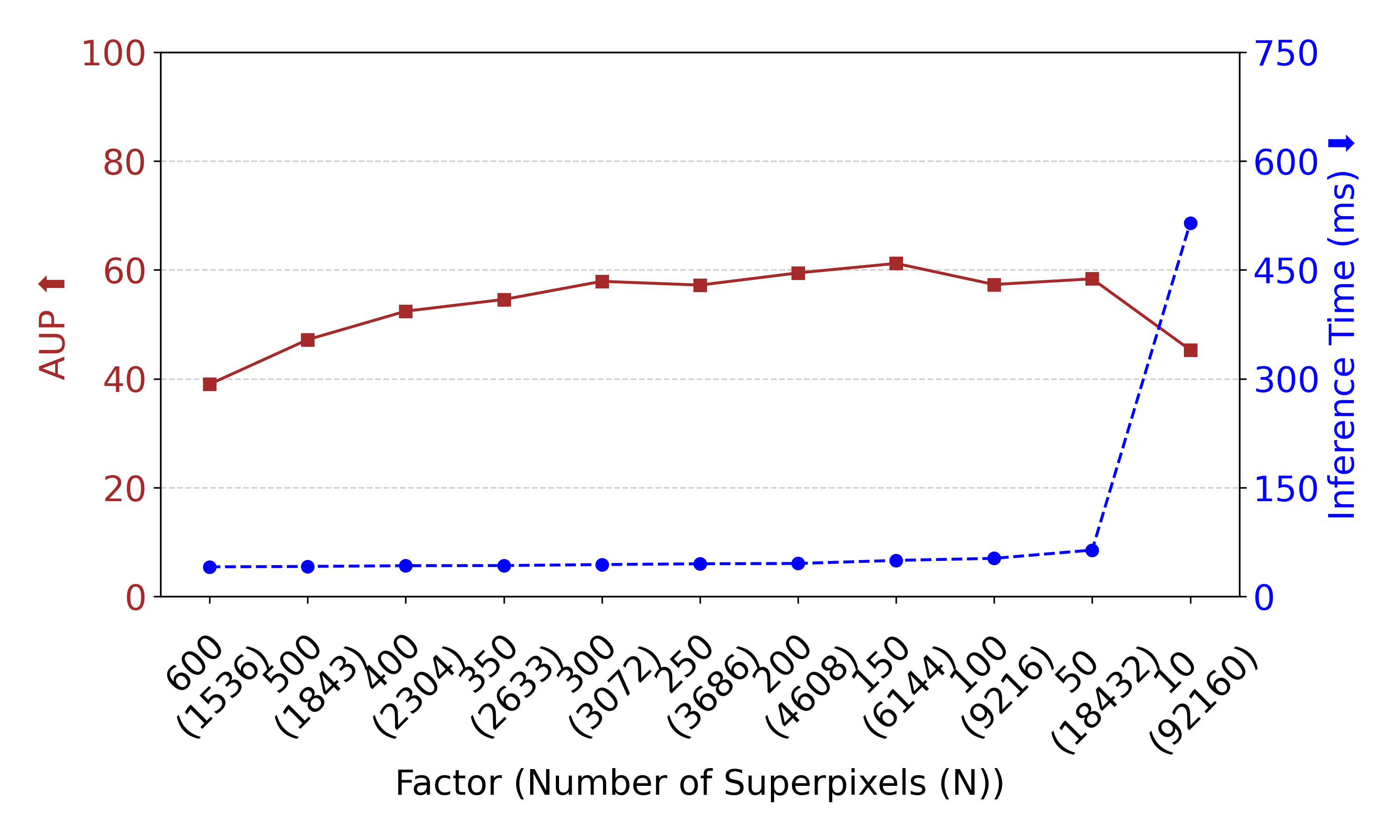}
     \vspace{-0.05in}
\caption{Variation of OOD detection performance and inference speed with the number of superpixels $N$. Averaged over six random runs on SMIYC-Anomaly.}
  \label{fig:no_sup_analysis}
   \vspace{-0.1in}
\end{figure}

\begin{figure*}[tb!]
    \centering
    \setlength{\tabcolsep}{1pt}
   \begin{tabular}{@{}ccccc@{}}
        \begin{subfigure}{0.2\textwidth}
            \includegraphics[width=\linewidth]{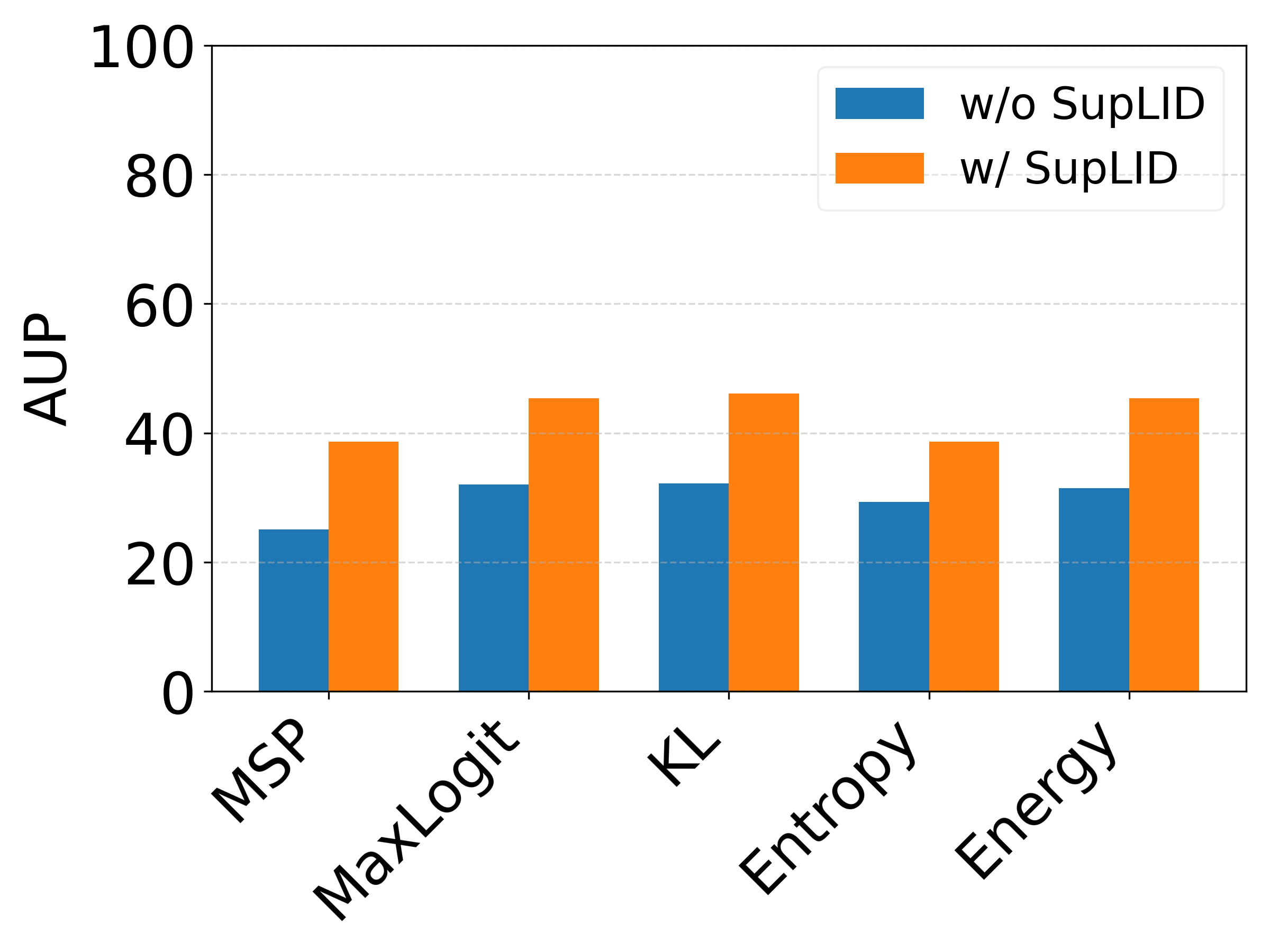}
            \caption{Baseline (DeepLabv3+)}
        \end{subfigure} &
        \begin{subfigure}{0.2\textwidth}
            \includegraphics[width=\linewidth]{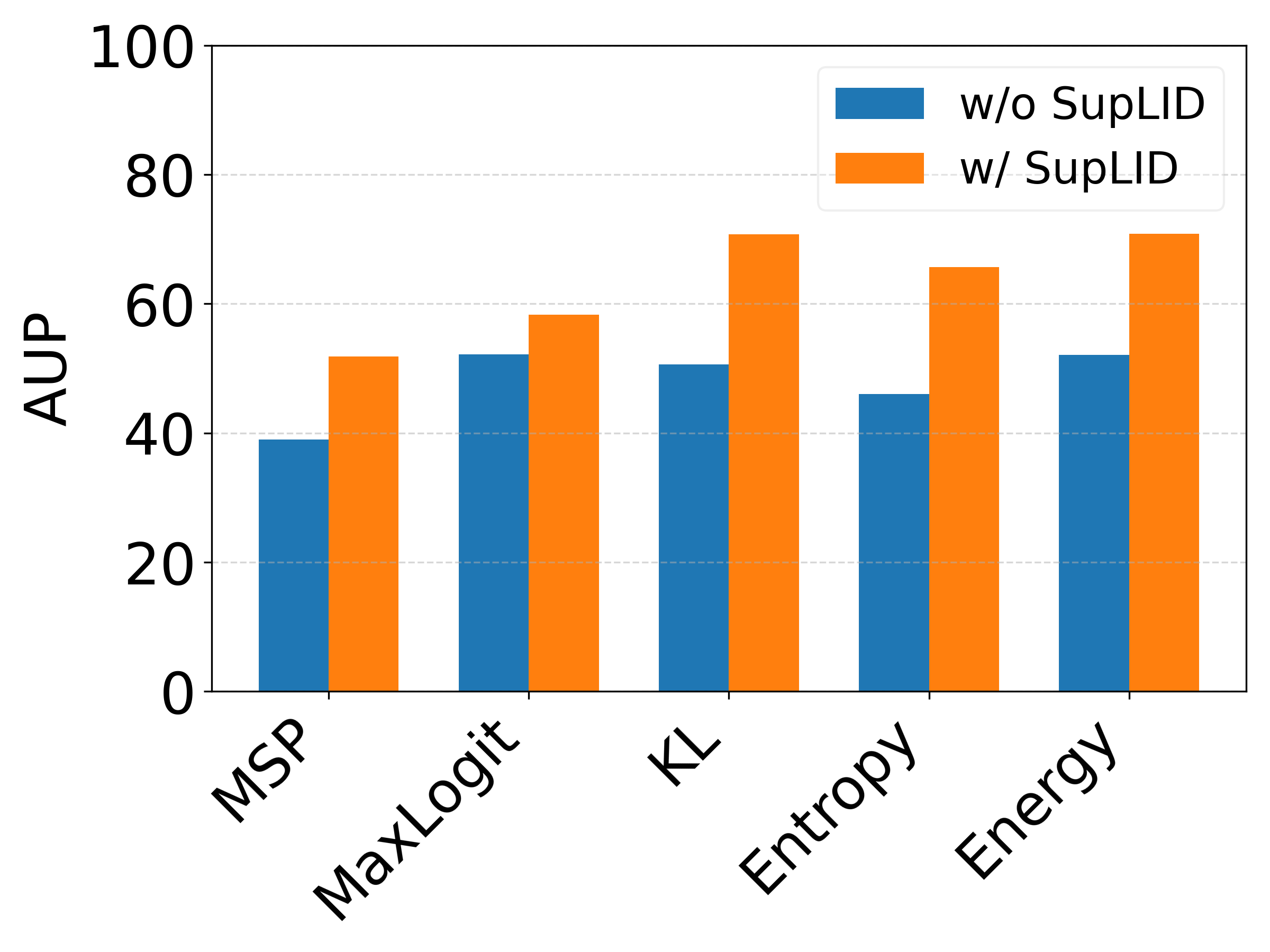}
            \caption{PEBAL~\cite{PEBAL}}
        \end{subfigure} &
        \begin{subfigure}{0.2\textwidth}
            \includegraphics[width=\linewidth]{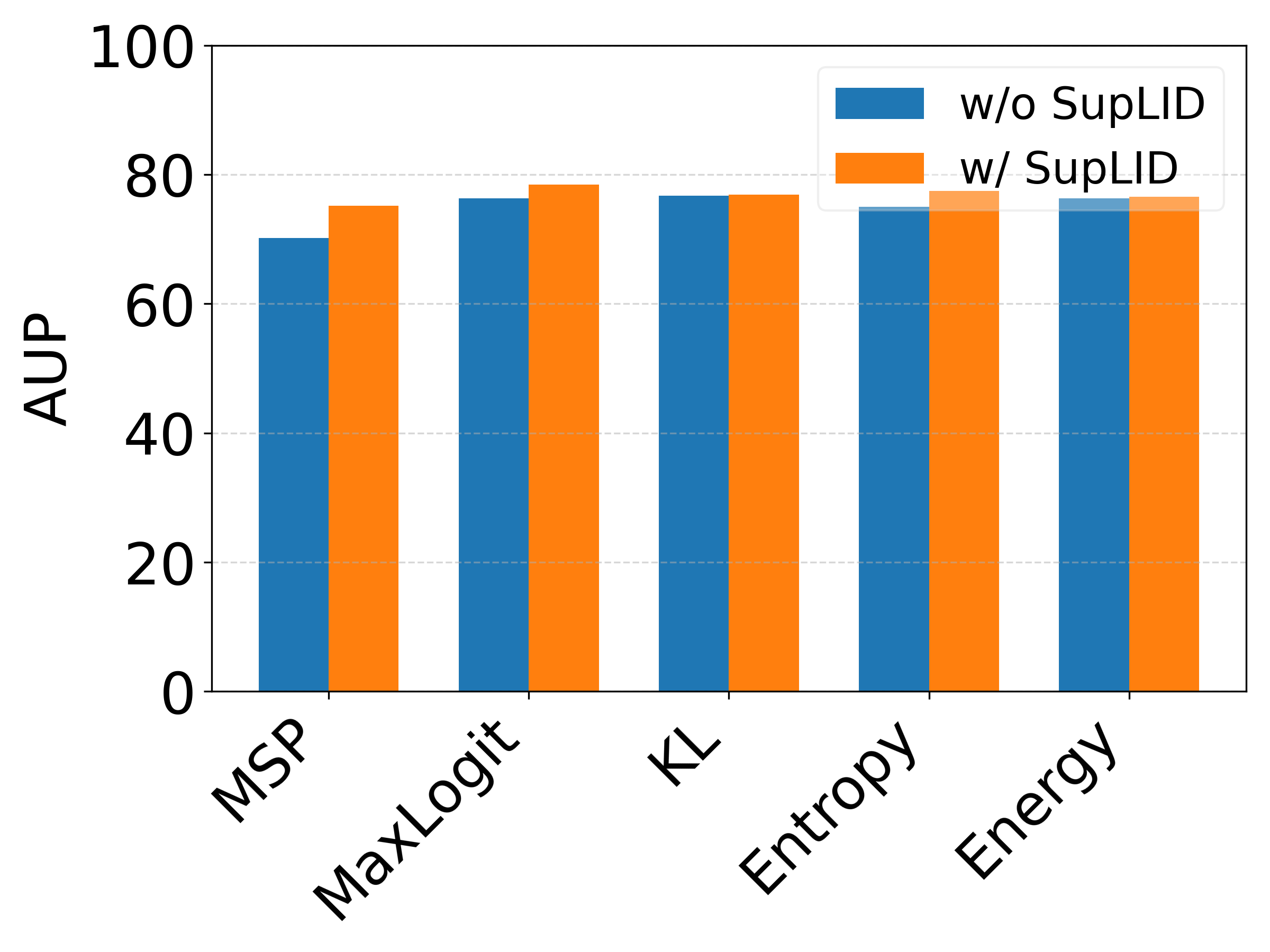}
            \caption{MetaOOD~\cite{MetaOOD}}
        \end{subfigure} &
        \begin{subfigure}{0.2\textwidth}
            \includegraphics[width=\linewidth]{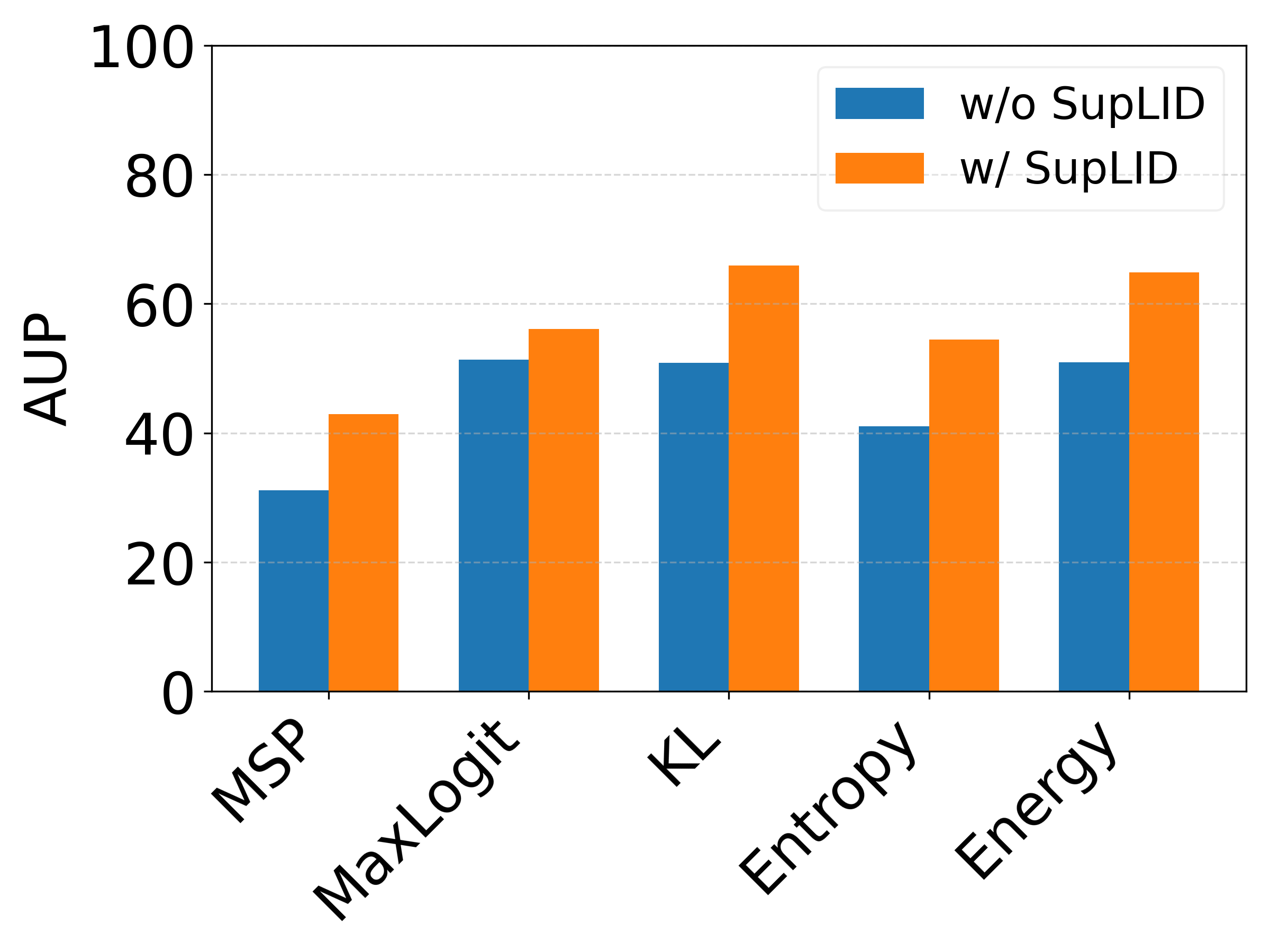}
            \caption{BE-OE~\cite{choiBalancedEnOE}}
        \end{subfigure} &
        \begin{subfigure}{0.2\textwidth}
            \includegraphics[width=\linewidth]{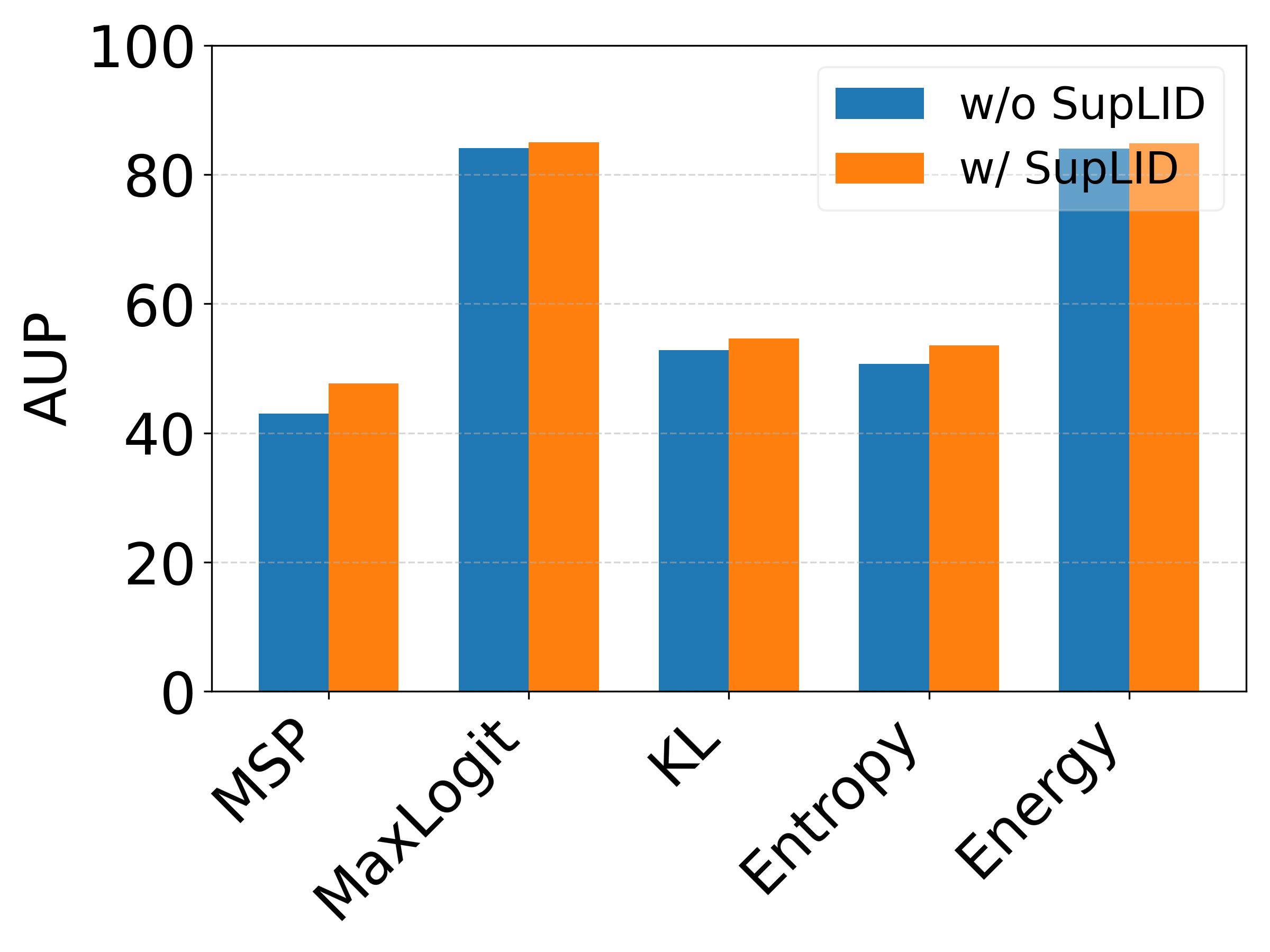}
            \caption{RPL~\cite{RPL}}
        \end{subfigure} \\
    \end{tabular}
     \vspace{-0.1in}
    \caption{Compatibility of SupLID with various classifier-based confidence scores across different baselines. The average performance across five OOD datasets is reported.}
    \label{fig:bar_plot_all_datasets}
\end{figure*}

\begin{table}[tb!]
\centering
\caption{Inference speed (averaged per-image) of the proposed SupLID when inferring with NVIDIA-A100 GPU.}
\begin{small}
\setlength{\tabcolsep}{4pt}
\begin{tabular}{l c}
\toprule
\textbf{Method} & \textbf{Inference Time (ms)} \\
\midrule
PEBAL~\cite{PEBAL} & 32.93 \\
RPL~\cite{RPL} & 31.14 \\
kNN~\cite{KNN_OOD} & 43.61 \\
NNGuide~\cite{NNGuide} & 43.69 \\
SynBoost~\cite{synboost} & 988.83 \\
SupLID (Ours) & {45.65} \\
\bottomrule
\end{tabular}
\end{small}
\label{tab:inference_time}
 \vspace{-0.2in}
\end{table}

\begin{table*}[tb!]
\centering
\caption{Performance comparison of SupLID with various classifier-based confidence scores across all models. Each classifier-based score is shown as \textbf{w/o SupLID / w/ SupLID}, with bold indicating the better value (lower FPR, higher AUP/AUR).}
\resizebox{\textwidth}{!}{%
\begin{tabular}{ll*{18}{c}}
\toprule
\textbf{} & \textbf{Method} & \multicolumn{3}{c}{\textbf{SMIYC-Anomaly}} & \multicolumn{3}{c}{\textbf{SMIYC-Obstacle}} & \multicolumn{3}{c}{\textbf{RoadAnomaly}} & \multicolumn{3}{c}{\textbf{Fishyscapes-Static}} & \multicolumn{3}{c}{\textbf{Fishyscapes-L\&F}} & \multicolumn{3}{c}{\textbf{Average}} \\
\cmidrule(lr){3-5} \cmidrule(lr){6-8} \cmidrule(lr){9-11} \cmidrule(lr){12-14} \cmidrule(lr){15-17} \cmidrule(lr){18-20}
& & FPR & AUP & AUR & FPR & AUP & AUR & FPR & AUP & AUR & FPR & AUP & AUR & FPR & AUP & AUR & FPR & AUP & AUR \\
\midrule
\multirow{7}{*}{\rotatebox{90}{Baseline}} & MSP & 60.33/\textbf{43.27} & 40.81/\textbf{57.59} & 78.90/\textbf{87.52} & \textbf{3.86}/8.06 & 41.24/\textbf{66.70} & \textbf{99.10}/98.19 & 71.24/\textbf{59.71} & 15.79/\textbf{22.71} & 67.60/\textbf{77.16} & \textbf{24.33}/37.63 & 22.51/\textbf{33.86} & \textbf{92.16}/90.98 & \textbf{40.25}/62.96 & 4.96/\textbf{12.62} & \textbf{89.39}/86.93 & \textbf{40.00}/42.33 & 25.06/\textbf{38.70} & 85.43/\textbf{88.16} \\
 & MaxLogit & 67.54/\textbf{44.83} & 45.99/\textbf{58.80} & 80.52/\textbf{87.88} & 4.48/\textbf{2.30} & 39.73/\textbf{69.45} & 98.72/\textbf{99.37} & 70.30/\textbf{52.85} & 19.05/\textbf{27.48} & 72.87/\textbf{81.16} & 18.17/\textbf{17.42} & 40.46/\textbf{46.36} & 95.73/\textbf{96.40} & \textbf{42.21}/50.33 & 14.93/\textbf{24.84} & \textbf{93.46}/92.27 & 40.54/\textbf{33.54} & 32.03/\textbf{45.39} & 88.26/\textbf{91.41} \\
 & KL & 66.77/\textbf{43.62} & 45.24/\textbf{59.33} & 80.68/\textbf{87.99} & 4.63/\textbf{2.28} & 34.51/\textbf{68.96} & 98.64/\textbf{99.36} & 69.65/\textbf{51.86} & 19.63/\textbf{30.31} & 73.51/\textbf{82.04} & \textbf{16.65}/16.76 & 43.20/\textbf{44.65} & 96.12/\textbf{96.44} & \textbf{39.78}/49.16 & 18.45/\textbf{27.11} & \textbf{94.08}/92.38 & 39.50/\textbf{32.74} & 32.21/\textbf{46.07} & 88.61/\textbf{91.64} \\
 & Entropy & 60.54/\textbf{58.50} & 45.73/\textbf{52.64} & 80.19/\textbf{82.54} & 3.85/\textbf{3.53} & 44.71/\textbf{57.17} & 99.09/\textbf{99.24} & 70.94/\textbf{67.31} & 17.01/\textbf{20.32} & 68.88/\textbf{71.92} & \textbf{23.60}/26.74 & 28.66/\textbf{42.72} & 93.08/\textbf{93.82} & \textbf{40.03}/42.37 & 10.62/\textbf{20.48} & 90.93/\textbf{92.15} & 39.79/\textbf{39.69} & 29.35/\textbf{38.67} & 86.44/\textbf{87.93} \\
 & Energy & 67.46/\textbf{44.55} & 45.44/\textbf{59.46} & 80.48/\textbf{87.94} & 4.49/\textbf{2.27} & 34.47/\textbf{68.96} & 98.67/\textbf{99.36} & 70.16/\textbf{52.52} & 19.54/\textbf{30.34} & 73.35/\textbf{81.98} & 17.77/\textbf{17.20} & 41.68/\textbf{43.32} & 95.90/\textbf{96.37} & \textbf{41.79}/50.07 & 16.05/\textbf{24.91} & \textbf{93.71}/92.12 & 40.34/\textbf{33.32} & 31.44/\textbf{45.40} & 88.43/\textbf{91.55} \\
 & kNN & 69.41 & 39.42 & 77.15 & 63.75 & 0.95 & 65.35 & 82.24 & 16.98 & 65.31 & 35.50 & 8.28 & 87.31 & 21.60 & 2.47 & 92.65 & 54.50 & 13.62 & 77.56 \\
 & NNGuide & 53.37 & 49.46 & 84.03 & 18.37 & 11.01 & 95.41 & 71.84 & 19.10 & 71.78 & 19.97 & 31.61 & 94.66 & 30.15 & 10.66 & 94.53 & 38.74 & 24.37 & 88.08 \\
\midrule
\multirow{7}{*}{\rotatebox{90}{PEBAL}} & MSP & 41.63/\textbf{31.39} & 48.70/\textbf{60.02} & 87.36/\textbf{91.11} & 6.75/\textbf{5.29} & 15.28/\textbf{35.03} & 97.58/\textbf{98.39} & 56.98/\textbf{41.16} & 35.20/\textbf{43.94} & 81.63/\textbf{86.59} & \textbf{14.16}/28.68 & 64.28/\textbf{70.42} & \textbf{96.93}/95.62 & \textbf{20.08}/32.66 & 31.47/\textbf{49.76} & 95.58/\textbf{95.99} & 27.92/\textbf{27.84} & 38.99/\textbf{51.83} & 91.82/\textbf{93.54} \\
 & MaxLogit & 36.19/\textbf{28.31} & 53.74/\textbf{62.15} & 89.35/\textbf{91.98} & 7.18/\textbf{6.08} & 13.08/\textbf{24.84} & 97.23/\textbf{97.74} & 45.26/\textbf{35.58} & 42.86/\textbf{51.60} & 87.10/\textbf{90.30} & \textbf{1.54}/3.06 & \textbf{92.06}/88.85 & \textbf{99.60}/99.34 & \textbf{5.08}/6.33 & 58.94/\textbf{63.94} & \textbf{98.94}/98.82 & 19.05/\textbf{15.87} & 52.14/\textbf{58.28} & 94.44/\textbf{95.64} \\
 & KL & 38.52/\textbf{29.00} & 47.55/\textbf{66.38} & 88.19/\textbf{92.05} & 7.83/\textbf{6.09} & 8.76/\textbf{78.68} & 96.35/\textbf{98.50} & 44.49/\textbf{34.73} & 41.98/\textbf{56.50} & 87.45/\textbf{91.22} & \textbf{1.44}/3.11 & \textbf{92.39}/86.61 & \textbf{99.65}/99.33 & \textbf{4.30}/5.51 & 62.29/\textbf{65.60} & \textbf{99.11}/98.93 & 19.32/\textbf{15.69} & 50.60/\textbf{70.75} & 94.15/\textbf{96.01} \\
 & Entropy & 41.61/\textbf{37.98} & 50.13/\textbf{63.16} & 88.45/\textbf{90.77} & 6.77/\textbf{5.60} & 10.83/\textbf{76.32} & 97.05/\textbf{98.81} & 55.71/\textbf{51.47} & 40.03/\textbf{49.64} & 84.65/\textbf{87.41} & \textbf{12.99}/13.15 & 78.77/\textbf{84.26} & 97.91/\textbf{98.24} & \textbf{18.87}/19.41 & 50.19/\textbf{55.14} & 96.92/\textbf{97.35} & 27.19/\textbf{25.52} & 45.99/\textbf{65.70} & 92.99/\textbf{94.52} \\
 & Energy & 36.49/\textbf{28.14} & 53.81/\textbf{69.13} & 89.07/\textbf{92.51} & 7.92/\textbf{6.74} & 10.45/\textbf{77.68} & 96.72/\textbf{98.39} & 44.58/\textbf{35.80} & 45.12/\textbf{57.40} & 87.63/\textbf{91.17} & \textbf{1.52}/3.15 & \textbf{92.09}/86.78 & \textbf{99.60}/99.30 & \textbf{4.76}/6.45 & 58.82/\textbf{63.12} & \textbf{98.96}/98.78 & 19.05/\textbf{16.06} & 52.06/\textbf{70.82} & 94.40/\textbf{96.03} \\
 & kNN & 61.40 & 39.56 & 79.06 & 74.86 & 0.86 & 61.37 & 84.25 & 18.04 & 65.30 & 36.03 & 10.81 & 88.48 & 22.01 & 3.06 & 93.35 & 55.71 & 14.47 & 77.51 \\
 & NNGuide & 29.41 & 54.44 & 90.01 & 15.71 & 5.02 & 93.11 & 48.71 & 40.85 & 86.06 & 1.62 & 92.69 & 99.55 & 4.71 & 51.96 & 98.98 & 20.03 & 48.99 & 93.54 \\
\midrule
\multirow{7}{*}{\rotatebox{90}{BE-OE}} & MSP & 53.16/\textbf{37.49} & 42.93/\textbf{56.00} & 82.19/\textbf{88.53} & 8.05/\textbf{4.01} & 11.55/\textbf{18.85} & 96.99/\textbf{98.25} & 56.27/\textbf{47.46} & 27.61/\textbf{37.34} & 79.55/\textbf{85.08} & \textbf{16.11}/32.11 & 48.88/\textbf{58.19} & \textbf{96.07}/94.68 & \textbf{20.39}/41.46 & 24.82/\textbf{44.01} & 94.72/\textbf{94.76} & \textbf{30.80}/32.51 & 31.16/\textbf{42.88} & 89.90/\textbf{92.26} \\
 & MaxLogit & 59.44/\textbf{39.35} & 45.97/\textbf{56.79} & 83.90/\textbf{88.96} & 10.11/\textbf{5.51} & 9.75/\textbf{18.12} & 96.30/\textbf{97.51} & 41.75/\textbf{36.24} & 40.62/\textbf{50.16} & 88.04/\textbf{90.81} & \textbf{1.23}/3.82 & \textbf{92.59}/86.61 & \textbf{99.55}/99.09 & \textbf{3.22}/8.33 & 67.80/\textbf{68.83} & \textbf{98.99}/98.43 & 23.15/\textbf{18.65} & 51.35/\textbf{56.10} & 93.36/\textbf{94.96} \\
 & KL & 58.53/\textbf{38.41} & 44.18/\textbf{59.22} & 83.89/\textbf{89.24} & 9.78/\textbf{5.39} & 8.32/\textbf{64.57} & 95.98/\textbf{98.26} & 40.77/\textbf{34.95} & 40.16/\textbf{53.19} & 88.27/\textbf{91.22} & \textbf{1.47}/4.12 & \textbf{92.18}/84.11 & \textbf{99.62}/99.08 & \textbf{2.91}/6.73 & \textbf{69.39}/68.46 & \textbf{99.10}/98.59 & 22.69/\textbf{17.92} & 50.85/\textbf{65.91} & 93.37/\textbf{95.28} \\
 & Entropy & 53.62/\textbf{50.63} & 45.24/\textbf{54.38} & 83.70/\textbf{86.16} & 8.18/\textbf{5.44} & 9.98/\textbf{42.86} & 96.68/\textbf{98.22} & 55.04/\textbf{51.46} & 34.29/\textbf{42.84} & 83.13/\textbf{86.02} & \textbf{15.17}/17.24 & 70.18/\textbf{79.41} & 97.29/\textbf{97.72} & \textbf{19.42}/20.44 & 45.34/\textbf{52.84} & 96.18/\textbf{96.81} & 30.29/\textbf{29.04} & 41.01/\textbf{54.47} & 91.40/\textbf{92.99} \\
 & Energy & 58.95/\textbf{39.19} & 45.28/\textbf{59.14} & 83.68/\textbf{89.03} & 11.03/\textbf{6.06} & 8.31/\textbf{60.32} & 95.76/\textbf{98.01} & 41.46/\textbf{36.69} & 41.49/\textbf{53.54} & 88.31/\textbf{91.10} & \textbf{1.17}/4.01 & \textbf{92.49}/84.36 & \textbf{99.55}/99.04 & \textbf{2.94}/8.77 & \textbf{67.07}/66.94 & \textbf{99.03}/98.39 & 23.11/\textbf{18.95} & 50.93/\textbf{64.86} & 93.27/\textbf{95.12} \\
 & kNN & 63.15 & 38.17 & 78.59 & 70.76 & 0.84 & 61.66 & 83.26 & 18.40 & 66.52 & 33.45 & 11.26 & 89.06 & 21.54 & 2.80 & 93.21 & 54.43 & 14.29 & 77.81 \\
 & NNGuide & 52.60 & 43.78 & 83.93 & 17.97 & 4.28 & 92.16 & 45.67 & 40.82 & 87.47 & 0.99 & 94.12 & 99.54 & 3.78 & 63.41 & 99.19 & 24.20 & 49.28 & 92.46 \\
\midrule
\multirow{7}{*}{\rotatebox{90}{MetaOOD}} & MSP & 14.12/\textbf{11.67} & 83.96/\textbf{85.80} & 96.68/\textbf{97.22} & \textbf{0.43}/0.49 & \textbf{94.23}/94.17 & \textbf{99.91}/99.88 & 12.34/\textbf{11.24} & 82.86/\textbf{84.11} & 97.39/\textbf{97.60} & 15.31/\textbf{15.16} & 60.73/\textbf{73.27} & 96.57/\textbf{97.33} & \textbf{38.19}/45.88 & 28.93/\textbf{38.31} & 91.57/\textbf{91.88} & \textbf{16.08}/16.89 & 70.14/\textbf{75.13} & 96.42/\textbf{96.78} \\
 & MaxLogit & 23.38/\textbf{18.58} & 81.35/\textbf{83.41} & 95.23/\textbf{96.10} & \textbf{0.45}/0.59 & \textbf{94.08}/93.63 & \textbf{99.91}/99.85 & 10.88/\textbf{10.06} & 83.96/\textbf{85.02} & 97.63/\textbf{97.79} & 4.40/\textbf{3.94} & 80.41/\textbf{84.30} & 98.96/\textbf{99.10} & 30.03/\textbf{29.61} & 41.67/\textbf{45.79} & 95.56/\textbf{95.57} & 13.83/\textbf{12.56} & 76.29/\textbf{78.43} & 97.46/\textbf{97.68} \\
 & KL & 27.71/\textbf{20.45} & 77.81/\textbf{80.01} & 94.29/\textbf{95.56} & \textbf{0.44}/0.69 & \textbf{94.10}/92.99 & \textbf{99.91}/99.83 & 10.49/\textbf{9.77} & \textbf{85.52}/80.93 & \textbf{97.80}/97.69 & 3.90/\textbf{3.77} & 81.09/\textbf{83.75} & 99.05/\textbf{99.13} & \textbf{28.46}/28.47 & 44.94/\textbf{46.52} & \textbf{95.95}/95.78 & 14.20/\textbf{12.63} & 76.69/\textbf{76.84} & 97.40/\textbf{97.60} \\
 & Entropy & 17.46/\textbf{15.45} & 80.76/\textbf{82.13} & 95.91/\textbf{96.43} & \textbf{0.43}/0.48 & 94.22/\textbf{94.50} & \textbf{99.90}/\textbf{99.90} & 10.85/\textbf{10.04} & \textbf{85.19}/81.91 & 97.72/\textbf{97.76} & 13.98/\textbf{13.48} & 73.39/\textbf{80.86} & 97.56/\textbf{98.04} & 37.40/\textbf{35.98} & 41.36/\textbf{47.71} & 93.11/\textbf{93.68} & 16.02/\textbf{15.09} & 74.98/\textbf{77.42} & 96.84/\textbf{97.16} \\
 & Energy & 28.10/\textbf{20.68} & 77.48/\textbf{79.92} & 94.24/\textbf{95.53} & \textbf{0.47}/0.72 & \textbf{93.87}/92.83 & \textbf{99.90}/99.83 & 10.48/\textbf{9.75} & \textbf{85.23}/81.12 & \textbf{97.79}/97.70 & 4.03/\textbf{3.81} & 81.13/\textbf{83.47} & 99.02/\textbf{99.11} & 29.14/\textbf{28.91} & 43.90/\textbf{45.51} & \textbf{95.84}/95.70 & 14.44/\textbf{12.78} & 76.32/\textbf{76.57} & 97.36/\textbf{97.57} \\
 & kNN & 40.19 & 40.78 & 83.73 & 28.95 & 4.46 & 88.59 & 47.84 & 21.11 & 78.42 & 45.62 & 5.51 & 82.79 & 34.81 & 1.29 & 87.72 & 39.48 & 14.63 & 84.25 \\
 & NNGuide & 19.14 & 77.42 & 95.24 & 0.79 & 92.11 & 99.79 & 12.76 & 81.19 & 97.28 & 8.88 & 73.10 & 98.16 & 36.85 & 22.60 & 93.68 & 15.68 & 69.28 & 96.83 \\
\midrule
\multirow{7}{*}{\rotatebox{90}{RPL}} & MSP & \textbf{35.33}/39.16 & 59.68/\textbf{62.14} & 89.40/\textbf{89.65} & 3.23/\textbf{2.99} & \textbf{31.44}/31.42 & \textbf{98.91}/98.85 & \textbf{51.51}/54.67 & 30.24/\textbf{32.15} & 80.53/\textbf{80.89} & \textbf{7.28}/9.51 & 62.35/\textbf{76.11} & 98.08/\textbf{98.10} & \textbf{10.44}/13.90 & 31.42/\textbf{36.40} & \textbf{97.43}/97.06 & \textbf{21.56}/24.05 & 43.02/\textbf{47.64} & 92.87/\textbf{92.91} \\
 & MaxLogit & 7.94/\textbf{7.25} & 88.14/\textbf{89.33} & 97.92/\textbf{98.11} & \textbf{0.11}/0.12 & \textbf{96.46}/96.22 & \textbf{99.96}/99.81 & 19.46/\textbf{18.48} & 70.01/\textbf{71.87} & 95.35/\textbf{95.66} & 0.82/\textbf{0.59} & 93.05/\textbf{94.64} & 99.74/\textbf{99.81} & \textbf{2.48}/3.94 & 72.65/\textbf{72.70} & \textbf{99.39}/99.18 & 6.16/\textbf{6.08} & 84.06/\textbf{84.95} & 98.47/\textbf{98.51} \\
 & KL & 23.15/\textbf{21.75} & 66.95/\textbf{68.90} & 93.22/\textbf{93.66} & 2.98/\textbf{2.76} & \textbf{19.41}/18.61 & \textbf{98.65}/98.44 & 38.71/\textbf{36.87} & 36.05/\textbf{37.69} & 87.00/\textbf{87.75} & 1.14/\textbf{0.63} & 91.17/\textbf{94.79} & 99.65/\textbf{99.78} & \textbf{2.35}/3.40 & 50.38/\textbf{53.11} & \textbf{99.33}/99.12 & 13.67/\textbf{13.09} & 52.79/\textbf{54.62} & 95.57/\textbf{95.75} \\
 & Entropy & 33.75/\textbf{33.46} & 65.27/\textbf{67.19} & 91.28/\textbf{91.77} & 3.06/\textbf{2.72} & \textbf{26.07}/25.66 & 98.91/\textbf{98.94} & 50.10/\textbf{49.14} & 34.92/\textbf{36.40} & 83.55/\textbf{84.24} & 5.76/\textbf{3.35} & 80.43/\textbf{88.22} & 98.97/\textbf{99.31} & 8.89/\textbf{8.55} & 46.81/\textbf{50.22} & 98.48/\textbf{98.65} & 20.31/\textbf{19.44} & 50.70/\textbf{53.54} & 94.24/\textbf{94.58} \\
 & Energy & 7.18/\textbf{6.74} & 88.55/\textbf{89.77} & 98.06/\textbf{98.24} & \textbf{0.09}/0.10 & \textbf{96.91}/96.42 & \textbf{99.96}/99.81 & 17.74/\textbf{17.48} & 71.60/\textbf{73.03} & 95.72/\textbf{95.90} & 0.85/\textbf{0.62} & 92.46/\textbf{94.16} & 99.73/\textbf{99.80} & \textbf{2.52}/4.11 & 70.61/\textbf{70.85} & \textbf{99.39}/99.16 & \textbf{5.68}/5.81 & 84.03/\textbf{84.85} & 98.57/\textbf{98.58} \\
 & kNN & 38.99 & 30.16 & 81.21 & 5.92 & 15.33 & 97.62 & 44.14 & 21.92 & 81.21 & 17.67 & 8.77 & 90.28 & 33.67 & 0.88 & 83.41 & 28.08 & 15.41 & 86.75 \\
 & NNGuide & 6.40 & 89.73 & 98.25 & 0.10 & 96.40 & 99.84 & 17.62 & 72.91 & 95.89 & 0.63 & 94.07 & 99.77 & 4.45 & 70.90 & 98.95 & 5.84 & 84.80 & 98.54 \\

\bottomrule
\end{tabular}%
}
\label{tab:ablation_on_other_scores}
\end{table*}

\subsubsection{Hyperparameter analysis} \label{sec:hyper_para} 
We evaluate the effect of hyperparameters of SupLID: the number of superpixels $N$, the number of neighborhood samples $k$, and the size of the ID coreset per class $m$.
In~\Cref{fig:k_analysis}, we show the effect of varying $k$ across several values of $m$ selected from \{10, 20, 50, 100, 200, 300, 400, 500, 600, 800, 1000, 2000\}.
As observed, the performance remains stable as long as both $k$ and $m$ are not too small (i.e., greater than 50).
As $m$ increases, the optimal performance also shifts toward higher values of $k$.
However, similar optimal performance can still be achieved with smaller coreset sizes (typically on the order of 100) for $m$, and consequently smaller values of $k$.

To study the effect of the number of superpixels, we determine $N$ by dividing the total number of pixels by a factor (which indicates the average number of pixels per segment), chosen from the set $\{10, 50, 100, 150, 200, 250, 300, 350, 400, 500\}$.
As shown in~\Cref{fig:no_sup_analysis}, the performance remains stable across varying values of $N$, except when it is set too high (approaching the original resolution) or too low (resulting in overly coarse segmentation with excessive smoothing).
We also indicate the average inference time across varying $N$ in the same graph.
As observed, when $N$ is within the range that yields optimal performance, varying $N$ does not significantly affect inference speed.

We also report the inference speed of the proposed SupLID score for the selected hyperparameters (in~\Cref{sec:experimental_setup}) in~\Cref{tab:inference_time}. Alongside SupLID, other post-hoc scores—kNN and NNGuide—are applied to the RPL model using the same coreset and $k$ value for a fair comparison.
Note that the coreset, based on the embedding vectors of training ID data, needs to be computed only once and is saved for use during inference.
As seen, the computational overhead added by our method is small, and significantly lower than that of hybrid approaches based on reconstruction (e.g., SynBoost~\cite{synboost}).

\section{Conclusion}
\looseness=-1
Developing robust OOD detection methods for semantic segmentation is crucial to prevent safety-critical hazards in applications such as autonomous driving.
We propose a simple yet effective post-hoc approach, SupLID, for OOD detection in semantic segmentation, which can be seamlessly integrated into segmentation models at deployment.
SupLID leverages the geometrical properties of the semantic space to guide classifier-based OOD scores of existing segmentation models.
It achieves dimensionality-aware, pixel-level OOD detection by incorporating a geometrical coreset to summarize the ID subspace and by aggregating scores at the superpixel level.
SupLID has been thoroughly evaluated on various segmentation models and diverse OOD datasets, effectively enhancing classifier-derived detection performance and achieving SOTA results.

\vspace{-0.1in}
\begin{acks}
This research used the LIEF HPC-GPGPU Facility at the University of Melbourne, established with LIEF Grant LE170100200.
Sarah Erfani is in part supported by Australian Research Council (ARC) Discovery Early Career Researcher Award (DECRA) DE220100680.
\end{acks}

\appendix

\section{Appendix}
\vspace{0.1in}
\subsection{Detailed Ablation}
We show detailed results of SupLID's compatibility with various classifier-based confidence scores in~\Cref{tab:ablation_on_other_scores}.
Additionally, we provide a summary of SupLID's performance across all OOD datasets and models using these confidence scores in~\Cref{fig:bar_plot_all_datasets}, which demonstrates that SupLID consistently improves the performance of all these confidence scores.

\subsection{Limitations}
SupLID currently leverages the geometrical properties of the semantic space to guide classifier-based confidence scores.
While a natural extension is to distance-based scoring functions, its performance as a post-hoc method remains constrained to some extent by the confidence estimates of the underlying segmentation model.
Nevertheless, as demonstrated in the main results, it remains applicable even to lower-performing models, thereby boosting their OOD detection performance.



\newpage
\noindent\textbf{GenAI Disclosure:}  We used ChatGPT (OpenAI, GPT-4 Turbo) solely for minor grammar and language corrections. All scientific content and analysis were entirely developed by the authors.

\bibliographystyle{ACM-Reference-Format}
\balance
\bibliography{sample-base}

@String{Computing = "Computing" }

@String{Computer = "{IEEE} Computer" }

@String{Springer = "Springer-Verlag" }

@ArtifactSoftware{R,
    title = {R: A Language and Environment for Statistical Computing},
    author = {{R Core Team}},
    organization = {R Foundation for Statistical Computing},
    address = {Vienna, Austria},
    year = {2019},
    url = {https://www.R-project.org/},
}

@String(PAMI  = {IEEE Trans. Pattern Anal. Mach. Intell.})

@String(IJCV  = {Int. J. Comput. Vis.})

@String(CVPR  = {IEEE Conf. Comput. Vis. Pattern Recog.})

@String(ICCV  = {Int. Conf. Comput. Vis.})

@String(ECCV  = {Eur. Conf. Comput. Vis.})

@String(NeurIPS = {Adv. Neural Inform. Process. Syst.})

@String(ICML  = {Int. Conf. Mach. Learn.})

@String(ICLR  = {Int. Conf. Learn. Represent.})

@String(ACCV  = {Asian Conf. Comput. Vis.})

@String(PAMI  = {IEEE TPAMI})

@String(IJCV  = {IJCV})

@String(CVPR  = {CVPR})

@String(ICCV  = {ICCV})

@String(ECCV  = {ECCV})

@String(NeurIPS = {NeurIPS})

@String(ICML  = {ICML})

@String(ICLR  = {ICLR})

@String(ACCV  = {ACCV})

@inproceedings{hendrycks_2017MSP,
  author       = {Dan Hendrycks and
                  Kevin Gimpel},
  title        = {A Baseline for Detecting Misclassified and Out-of-Distribution Examples
                  in Neural Networks},
  booktitle    = ICLR,
  year         = {2017},
}

@inproceedings{lee_2018_mahalanobis,
  title={A simple unified framework for detecting out-of-distribution samples and adversarial attacks},
  author={Lee, Kimin and Lee, Kibok and Lee, Honglak and Shin, Jinwoo},
  booktitle=NeurIPS,
  year={2018},
}

@inproceedings{choiBalancedEnOE,
  title={Balanced Energy Regularization Loss for Out-of-distribution Detection},
  author={Choi, Hyunjun and Jeong, Hawook and Choi, Jin Young},
  booktitle=CVPR,
  pages={15691--15700},
  year={2023}
}

@inproceedings{hendrycks_2018OE,
  author       = {Dan Hendrycks and
                  Mantas Mazeika and
                  Thomas G. Dietterich},
  title        = {Deep Anomaly Detection with Outlier Exposure},
  booktitle    = ICLR,
  year         = {2019},
}

@inproceedings{EnergyOE,
  title={Energy-based out-of-distribution detection},
  author={Liu, Weitang and Wang, Xiaoyun and Owens, John and Li, Yixuan},
  booktitle=NeurIPS,
  pages={21464--21475},
  year={2020}
}

@inproceedings{KNN_OOD,
  title={Out-of-distribution detection with deep nearest neighbors},
  author={Sun, Yiyou and Ming, Yifei and Zhu, Xiaojin and Li, Yixuan},
  booktitle=ICML,
  pages={20827--20840},
  year={2022},
  organization={PMLR}
}

@inproceedings{LID_reconstruction,
  title={A dimensionality-driven approach for unsupervised out-of-distribution detection},
  author={Wang, Qizhou and Erfani, Sarah M and Leckie, Christopher and Houle, Michael E},
  booktitle={Proceedings of the 2021 SIAM International Conference on Data Mining (SDM)},
  pages={118--126},
  year={2021},
  organization={SIAM}
}

@inproceedings{LID_OOD_correlation,
  title={On the correlation between local intrinsic dimensionality and outlierness},
  author={Houle, Michael E and Schubert, Erich and Zimek, Arthur},
  booktitle={Similarity Search and Applications: 11th International Conference, SISAP 2018, Lima, Peru, October 7--9, 2018, Proceedings 11},
  pages={177--191},
  year={2018},
  organization={Springer}
}

@inproceedings{DAO_LID_outlier,
  title={Dimensionality-Aware Outlier Detection},
  author={Anderberg, Alastair and Bailey, James and Campello, Ricardo JGB and Houle, Michael E and Marques, Henrique O and Radovanovi{\'c}, Milo{\v{s}} and Zimek, Arthur},
  booktitle={Proceedings of the 2024 SIAM International Conference on Data Mining (SDM)},
  pages={652--660},
  year={2024},
  organization={SIAM}
}

@inproceedings{PEBAL,
  title={Pixel-wise energy-biased abstention learning for anomaly segmentation on complex urban driving scenes},
  author={Tian, Yu and Liu, Yuyuan and Pang, Guansong and Liu, Fengbei and Chen, Yuanhong and Carneiro, Gustavo},
  booktitle=ECCV,
  pages={246--263},
  year={2022},
  organization={Springer}
}

@inproceedings{NNGuide,
  title={Nearest neighbor guidance for out-of-distribution detection},
  author={Park, Jaewoo and Jung, Yoon Gyo and Teoh, Andrew Beng Jin},
  booktitle=ICCV,
  pages={1686--1695},
  year={2023}
}

@inproceedings{LID_adversarial,
  title     = {Characterizing Adversarial Subspaces Using Local Intrinsic Dimensionality},
  author    = {Ma, Xingjun and Li, Bo and Wang, Yisen and Erfani, Sarah M. and Wijewickrema, Sudanthi and Schoenebeck, Grant and Song, Dawn and Houle, Michael E. and Bailey, James},
  booktitle = ICLR,
  year      = {2018},
  month     = {Apr},
  address   = {Vancouver, Canada},
}

@InProceedings{MaxLogit,
  title = 	 {Scaling Out-of-Distribution Detection for Real-World Settings},
  author =       {Hendrycks, Dan and Basart, Steven and Mazeika, Mantas and Zou, Andy and Kwon, Joseph and Mostajabi, Mohammadreza and Steinhardt, Jacob and Song, Dawn},
  booktitle = 	 ICML,
  pages = 	 {8759--8773},
  year = 	 {2022}
}

@inproceedings{MetaOOD,
  title={Entropy maximization and meta classification for out-of-distribution detection in semantic segmentation},
  author={Chan, Robin and Rottmann, Matthias and Gottschalk, Hanno},
  booktitle=ICCV,
  pages={5128--5137},
  year={2021}
}

@inproceedings{RPL,
  title={Residual pattern learning for pixel-wise out-of-distribution detection in semantic segmentation},
  author={Liu, Yuyuan and Ding, Choubo and Tian, Yu and Pang, Guansong and Belagiannis, Vasileios and Reid, Ian and Carneiro, Gustavo},
  booktitle=ICCV,
  pages={1151--1161},
  year={2023}
}

@article{Maxlogit_openset,
  title={Open-set recognition: A good closed-set classifier is all you need?},
  author={Vaze, Sagar and Han, Kai and Vedaldi, Andrea and Zisserman, Andrew},
  year={2021},
  publisher={OpenReview}
}

@inproceedings{standardMaxLogit,
  title={Standardized max logits: A simple yet effective approach for identifying unexpected road obstacles in urban-scene segmentation},
  author={Jung, Sanghun and Lee, Jungsoo and Gwak, Daehoon and Choi, Sungha and Choo, Jaegul},
  booktitle=ICCV,
  pages={15425--15434},
  year={2021}
}

@article{shannonEntropy,
  title={A mathematical theory of communication},
  author={Shannon, Claude E},
  journal={The Bell system technical journal},
  volume={27},
  number={3},
  pages={379--423},
  year={1948},
  publisher={Nokia Bell Labs}
}

@article{embeddingDensity,
  title = {Distance-based Confidence Score for Neural Network Classifiers},
  author = {Mandelbaum, Amit and Weinshall, Daphna},
  journal = {CoRR},
  volume = {abs/1709.09844},
  year = {2017},
}

@inproceedings{classicEx,
  title={Finding nearest neighbors in growth-restricted metrics},
  author={Karger, David R and Ruhl, Matthias},
  booktitle={Proceedings of the thiry-fourth annual ACM symposium on Theory of computing},
  pages={741--750},
  year={2002}
}

@inproceedings{GED,
  title={Generalized expansion dimension},
  author={Houle, Michael E and Kashima, Hisashi and Nett, Michael},
  booktitle={2012 IEEE 12th International Conference on Data Mining Workshops},
  pages={587--594},
  year={2012},
  organization={IEEE}
}

@inproceedings{LID,
  title={Local intrinsic dimensionality I: an extreme-value-theoretic foundation for similarity applications},
  author={Houle, Michael E},
  booktitle={Similarity Search and Applications: 10th International Conference, SISAP 2017, Munich, Germany, October 4-6, 2017, Proceedings 10},
  pages={64--79},
  year={2017},
  organization={Springer}
}

@inproceedings{estimatingLID,
  title={Estimating local intrinsic dimensionality},
  author={Amsaleg, Laurent and Chelly, Oussama and Furon, Teddy and Girard, St{\'e}phane and Houle, Michael E and Kawarabayashi, Ken-ichi and Nett, Michael},
  booktitle={Proceedings of the 21th ACM SIGKDD International Conference on Knowledge Discovery and Data Mining},
  pages={29--38},
  year={2015}
}

@article{SLIC,
  title={SLIC superpixels compared to state-of-the-art superpixel methods},
  author={Achanta, Radhakrishna and Shaji, Appu and Smith, Kevin and Lucchi, Aurelien and Fua, Pascal and S{\"u}sstrunk, Sabine},
  journal=PAMI,
  volume={34},
  number={11},
  pages={2274--2282},
  year={2012},
  publisher={IEEE}
}

@inproceedings{fishyscapes,
  title={Fishyscapes: A benchmark for safe semantic segmentation in autonomous driving},
  author={Blum, Hermann and Sarlin, Paul-Edouard and Nieto, Juan and Siegwart, Roland and Cadena, Cesar},
  booktitle={proceedings of the IEEE/CVF international conference on computer vision workshops},
  pages={0--0},
  year={2019}
}

@article{coreset_survey,
  title={Geometric approximation via coresets},
  author={Agarwal, Pankaj K and Har-Peled, Sariel and Varadarajan, Kasturi R and others},
  journal={Combinatorial and computational geometry},
  volume={52},
  number={1},
  pages={1--30},
  year={2005}
}

@inproceedings{COLLIDER_LID_coreset,
  title={COLLIDER: A robust training framework for backdoor data},
  author={Dolatabadi, Hadi Mohaghegh and Erfani, Sarah and Leckie, Christopher},
  booktitle=ACCV,
  pages={3893--3910},
  year={2022}
}

@inproceedings{cityscapes_dataset,
  title={The cityscapes dataset for semantic urban scene understanding},
  author={Cordts, Marius and Omran, Mohamed and Ramos, Sebastian and Rehfeld, Timo and Enzweiler, Markus and Benenson, Rodrigo and Franke, Uwe and Roth, Stefan and Schiele, Bernt},
  booktitle=CVPR,
  pages={3213--3223},
  year={2016}
}

@inproceedings{SMIYC_dataset,
	title={SegmentMeIfYouCan: A Benchmark for Anomaly Segmentation},
	author={Robin Chan and Krzysztof Lis and Svenja Uhlemeyer and Hermann Blum and Sina Honari and Roland Siegwart and Pascal Fua and Mathieu Salzmann and Matthias Rottmann},
	booktitle={Thirty-fifth Conference on Neural Information Processing Systems Datasets and Benchmarks Track},
	year={2021}
}

@inproceedings{RoadAnomaly_dataset,
  title={Detecting the unexpected via image resynthesis},
  author={Lis, Krzysztof and Nakka, Krishna and Fua, Pascal and Salzmann, Mathieu},
  booktitle=ICCV,
  pages={2152--2161},
  year={2019}
}

@article{FS_dataset,
  title={The fishyscapes benchmark: Measuring blind spots in semantic segmentation},
  author={Blum, Hermann and Sarlin, Paul-Edouard and Nieto, Juan and Siegwart, Roland and Cadena, Cesar},
  journal=IJCV,
  volume={129},
  number={11},
  pages={3119--3135},
  year={2021},
  publisher={Springer}
}

@inproceedings{DeepLab_NVIDIA,
  title={Improving semantic segmentation via video propagation and label relaxation},
  author={Zhu, Yi and Sapra, Karan and Reda, Fitsum A and Shih, Kevin J and Newsam, Shawn and Tao, Andrew and Catanzaro, Bryan},
  booktitle=CVPR,
  pages={8856--8865},
  year={2019}
}

@inproceedings{synboost,
  title={Pixel-wise anomaly detection in complex driving scenes},
  author={Di Biase, Giancarlo and Blum, Hermann and Siegwart, Roland and Cadena, Cesar},
  booktitle=CVPR,
  pages={16918--16927},
  year={2021}
}

@inproceedings{densehybrid,
  title={Densehybrid: Hybrid anomaly detection for dense open-set recognition},
  author={Grci{\'c}, Matej and Bevandi{\'c}, Petra and {\v{S}}egvi{\'c}, Sini{\v{s}}a},
  booktitle=ECCV,
  pages={500--517},
  year={2022},
  organization={Springer}
}

@inproceedings{diverse_coreset,
  title={Composable core-sets for diversity and coverage maximization},
  author={Indyk, Piotr and Mahabadi, Sepideh and Mahdian, Mohammad and Mirrokni, Vahab S},
  booktitle={Proceedings of the 33rd ACM SIGMOD-SIGACT-SIGART symposium on Principles of database systems},
  pages={100--108},
  year={2014}
}

@inproceedings{LID_noisy_label,
  title={Dimensionality-driven learning with noisy labels},
  author={Ma, Xingjun and Wang, Yisen and Houle, Michael E and Zhou, Shuo and Erfani, Sarah and Xia, Shutao and Wijewickrema, Sudanthi and Bailey, James},
  booktitle={ICML},
  pages={3355--3364},
  year={2018},
  organization={PMLR}
}

\end{document}